\documentclass[10pt,twocolumn,letterpaper]{article}

\usepackage[pagenumbers]{cvpr} 

\usepackage{graphicx}
\usepackage{booktabs}
\usepackage{multirow}
\usepackage[table]{xcolor}
\definecolor{mygray}{gray}{0.8}
\definecolor{lightCyan}{rgb}{0.95,0.95,0.95}
\usepackage{fix-cm}
\usepackage{amssymb}
\usepackage{bm}
\usepackage{pifont}
\usepackage{wrapfig}
\usepackage{xcolor}
\usepackage[accsupp]{axessibility} 

\definecolor{cvprblue}{rgb}{0.21,0.49,0.74}
\usepackage[pagebackref,breaklinks,colorlinks,allcolors=cvprblue]{hyperref}

\def\paperID{} 
\def\confName{CVPR}
\def\confYear{2026}

\title{Scaling Test-Time Robustness of Vision-Language Models via Self-Critical Inference Framework}

\author{Kaihua Tang\textsuperscript{1} \quad Jiaxin Qi\textsuperscript{2} \quad Jinli Ou\textsuperscript{4} \quad Yuhua Zheng\textsuperscript{3} \quad Jianqiang Huang\textsuperscript{2,3,4}\thanks{Corresponding author.}\\
\textsuperscript{1} Tongji University, China \quad \textsuperscript{2} Computer Network Information Center, CAS, China\\
\textsuperscript{3} HIAS, University of Chinese Academy of Sciences, China\\
\textsuperscript{4} University of Chinese Academy of Sciences, China \\
{\tt\small tangkaihua@tongji.edu.cn, jxqi@cnic.cn, oujinli@zuaa.zju.edu.cn}\\
{\tt\small zhengyuhua@ucas.ac.cn, jqhuang@cnic.cn}
}

\begin{document}
\maketitle

\begin{abstract}


The emergence of Large Language Models (LLMs) has driven rapid progress in multi-modal learning, particularly in the development of Large Vision-Language Models (LVLMs). However, existing LVLM training paradigms place excessive reliance on the LLM component, giving rise to two critical robustness challenges: language bias and language sensitivity. To address both issues simultaneously, we propose a novel Self-Critical Inference (SCI) framework that extends Visual Contrastive Decoding by conducting multi-round counterfactual reasoning through both textual and visual perturbations. This process further introduces a new strategy for improving robustness by scaling the number of counterfactual rounds. Moreover, we also observe that failure cases of LVLMs differ significantly across models, indicating that fixed robustness benchmarks may not be able to capture the true reliability of LVLMs. To this end, we propose the Dynamic Robustness Benchmark (DRBench), a model-specific evaluation framework targeting both language bias and sensitivity issues. Extensive experiments show that SCI consistently outperforms baseline methods on DRBench, and that increasing the number of inference rounds further boosts robustness beyond existing single-step counterfactual reasoning methods. \footnote{Our code is publicly available on GitHub: \url{https://github.com/KaihuaTang/Self-Critical-Inference-Framework}}


\end{abstract}

\section{Introduction}

The recent advance in Large Language Models~\citep{brown2020language,achiam2023gpt,touvron2023llama,bai2023qwen,wu2025generalization,liu2024deepseek} (LLMs) has not only revolutionized the field of natural language processing but also catalyzed significant progress in multi-modal research, particularly in the vision-language domain~\citep{yin2024survey,zhang2024mm,peng2025lmm}. To better utilize the knowledge of LLMs, the prevalent training framework for Large Vision-Language Model (LVLM) integrates a visual encoder with a pretrained LLM and jointly fine-tunes the combined architecture, resulting in powerful and versatility LVLMs such as InstructBLIP~\citep{dai2023instructblp}, LLaVA series~\citep{liu2023llava,liu2024llavanext} and Qwen-VL series~\citep{Qwen-VL,Qwen2-VL}.

\begin{figure*}[t]
    \centering
    \includegraphics[width=1.0\linewidth]{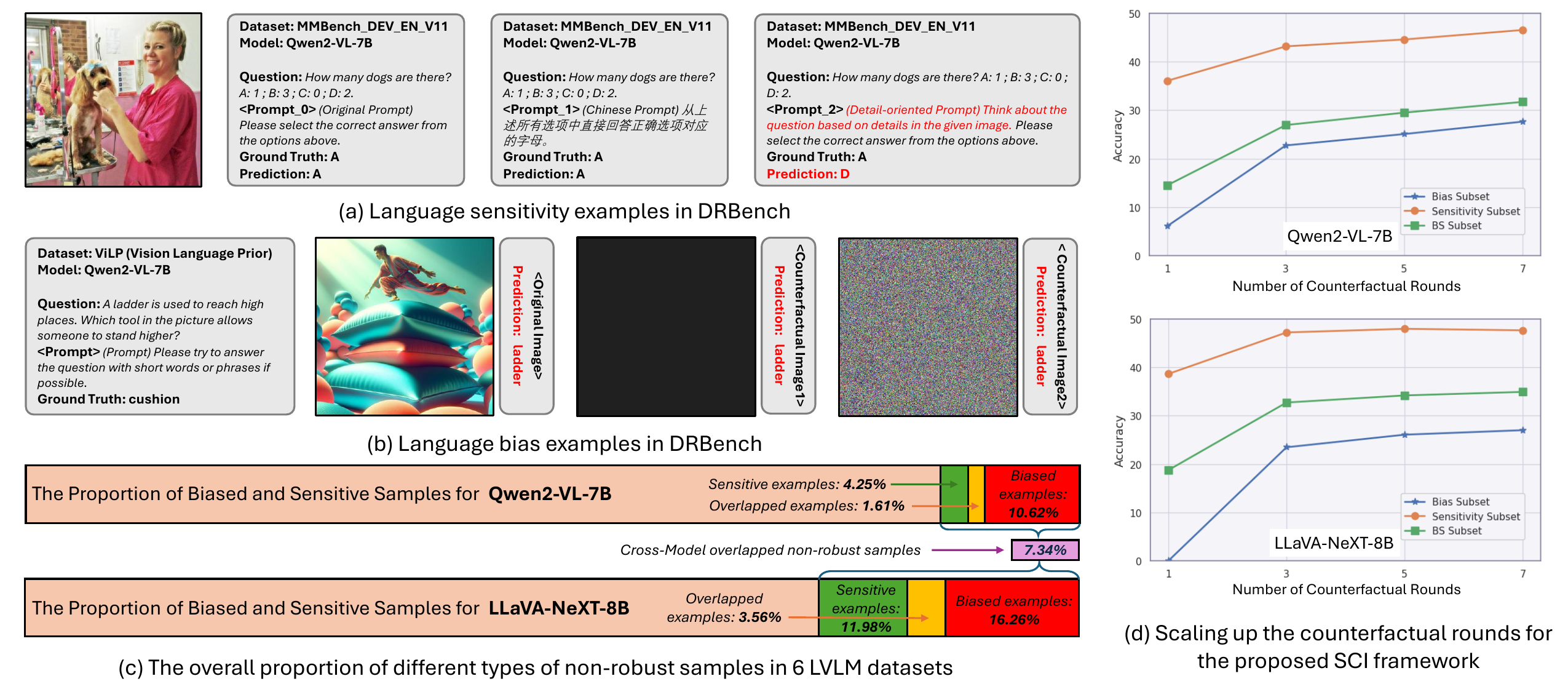}
    \caption{(a) and (b) are real DRBench examples suffering from language sensitivity and bias issues; (c) shows the overall proportion of different types of non-robust samples across all 6 datasets under two commonly used LVLMs; (d) demonstrates a novel test-time scaling strategy of robustness regarding the increased counterfactual rounds in the proposed SCI.}
    \label{fig:tf}
\end{figure*}

However, these LVLMs continue to suffer from robustness issues in two key aspects. First, the above-mentioned LLM-based vision-language framework inevitably inherits certain drawbacks of LLMs, such as sensitivity to language prompts~\citep{arora2023ask,jiang2023calibrating,wightman2023strength}. Conventional VQA models lack the large-scale pretraining of LLMs and thus can only understand very limited textual information, failing to capture subtle prompt variations and thereby side-stepping this issue. As illustrated in Figure~\ref{fig:tf}(a), simply requesting a LVLM to check image details without altering the question results in different outputs for the same input image. This language sensitivity undermines the consistency of LVLMs, reducing their reliability from the user's perspective. Second, vision-language models are also known to be susceptible to language bias. For example, conventional Visual Question Answering (VQA) models often rely heavily on language priors to answer questions, disregarding visual input~\citep{niu2021counterfactual,wen2021debiased}. As shown in Figure~\ref{fig:tf}(b), this problem also persists in LVLMs and can sometimes lead to generating non-existent contents, known as object hallucination~\citep{li2023evaluating,leng2024mitigating}.

Recently, a growing number of research has focused on mitigating object hallucination in LVLMs~\citep{zhou2024analyzing,li2023evaluating}. Among these efforts, Visual Contrastive Decoding (VCD)~\citep{leng2024mitigating} and its variants~\citep{woo2024don,suo2025octopus} have emerged as some of the most effective and widely adopted solutions. These methods typically perform a standard inference to obtain baseline logits and then estimate potential biases via a secondary inference with perturbed inputs. The final unbiased prediction is derived by weighted subtraction of the two logits. However, the object hallucination is merely a continuation of the language bias observed in conventional VLMs~\citep{niu2021counterfactual,tang2020unbiased}, and it ignores the issue of language sensitivity that is newly introduced by LVLMs.

In this work, we first analyze the underlying principles of VCD, particularly the role of the trade-off parameter $\alpha$, which is absent in the original Contrastive Decoding (CD)~\citep{li2023contrastive}. Through an in-depth mathematical analysis, we demonstrate that VCD is theoretically aligned with some debiasing algorithms used in previous vision-language tasks, such as TDE~\citep{tang2020unbiased} and TIE~\citep{niu2021counterfactual}. Specifically, VCD leverages TIE logits to reweight the original logits, where $1/\alpha$ acts as the temperature parameter for logit scaling. Building on this insight, we propose a more comprehensive inference paradigm, termed Self-Critical Inference \textbf{(SCI)} framework, which unifies both Textual Counterfactual (TC) and Visual Counterfactual (VC) components. The final prediction is then derived from aggregating and comparing all multi-round counterfactual logits. This approach generalizes VCD and enables the simultaneous mitigation of both bias and sensitivity issues. We further examine three configurations: SCI$_3$, SCI$_5$, and SCI$_7$, with different numbers of input variations to investigate the effect of increasing counterfactual inference rounds. We argue that our approach establishes a new potential test-time scaling direction, distinct from prior methods that increase intermediate context token lengths within a single inference. Instead, robustness can be enhanced by performing increased round of counterfactual inference.


We also introduce a new evaluation benchmark, termed Dynamic Robustness Benchmark \textbf{(DRBench)}, to adaptively assess the robustness improvements of individual models. The key motivation behind DRBench is that those non-robust data samples are not fixed. As shown in Figure~\ref{fig:tf}(c), among all 24.68\% hard samples for one LVLM (LLaVA-NeXT), there are only 7.34\% shared with another LVLM (Qwen2-VL). This suggests that an LVLM may perform perfectly well on a fixed robustness dataset, yet still be vulnerable to other new samples. Besides, to enable a more precise analysis of algorithmic contributions, it is essential to disentangle the robustness gains from the confounding effect of base model performance. To this end, this benchmark is constructed by adaptively extracting non-robust subsets from existing LVLM datasets, based on the performance of a given LVLM. These model-specific subsets prevent newly introduced LVLMs from covering robustness issues by overfitting to existing datasets. Notably, the DRBench is easily scalable and can be seamlessly applied to widely used real datasets such as MMBench, MME, \textit{etc.}, introducing more diverse and nature question types than previous datasets~\citep{li2023evaluating}. Furthermore, as illustrated in Figure~\ref{fig:tf}(c), the additional statistical information itself from DRBench facilitates a more comprehensive diagnosis of the inherent vulnerabilities of each LVLM.

The main contributions of this paper are threefold: 1) We propose SCI, a counterfactual inference framework that simultaneously mitigates language bias and enforces language consistency. 2) We introduce the DRBench, a model-specific and dynamic benchmark designed to better assess the robustness of LVLMs under samples from real downstream tasks. 3) We demonstrate that SCI consistently improves performance on both the DRBench and standard datasets, exhibiting strong generalizability. Furthermore, we reveal a previously underexplored potential for improving robustness by increasing the number of test-time counterfactual inference rounds.

\section{Related Work}

\noindent\textbf{Large vision-language models.} LVLMs integrate two of the most significant breakthroughs in recent years: the versatile image encoder CLIP~\citep{openai2021clip} and LLMs for general-purpose question answering~\citep{radford2019gpt2,touvron2023llama,peng2025lmm,wu2025generalization}. The typical inference pipeline of a LVLM proceeds as follows: the input image is first encoded by CLIP or its more advanced successors~\citep{zhai2023sigmoid} to extract patch-level visual features; an adapter then maps these features to the token embedding space of the LLM~\citep{liu2023llava,Qwen-VL}; finally, the visual and textual token embeddings are jointly fed into the LLM to generate the response. LVLMs have shown broad applicability in vision-language tasks such as image captioning~\citep{xu2015show,yang2019auto,yang2023exploring} and Visual Question Answering (VQA)~\citep{antol2015vqa}. 



\noindent\textbf{Language bias and sensitivity in vision-language models.} Language bias has been a longstanding challenge for visual-language models. Previously, it was widely studied as the language prior in tasks like VQA\cite{niu2021counterfactual,goyal2017making}. In today's LVLMs, it commonly manifests as object hallucination. Recent works have sought to mitigate it through targeted retraining and contrastive decoding strategies\cite{gunjal2024detecting,leng2024mitigating,jiang2025devils}, which are parallel to earlier techniques such as rebalanced training and counterfactual inference~\citep{chen2020counterfactual,niu2021counterfactual}. Meanwhile, sensitivity to language prompts has received considerably less attention in VL research. Early VQA systems side-stepping this issue by using a small language encoder. The emergence of LLMs has brought it to the forefront. Existing mitigation strategies can be broadly categorized into three groups: 1) prompt ensembling~\citep{pitis2023boosted}; 2) RL-based prompt optimization~\citep{kwon2024stableprompt}; 3) Chain-of-thought verification~\citep{wang2022self}.

\noindent\textbf{Test-time scaling.} Scaling laws have always been central to understanding LLM behavior, particularly the positive correlation between the scale of model/dataset/compute and the performance~\citep{kaplan2020scaling,hoffmann2022training}. Recently, the attention has shifted toward test-time scaling, where increasing inference-time compute is also critical~\citep{snell2025scaling}, such as adding demonstrations or decoding steps. In this work, we extend the notion of test-time scaling to the robustness: rather than increasing intermediate token length in a single inference, our method improves LVLM robustness by aggregating logits across more counterfactual inference rounds.

\section{Methodology}

\subsection{Preliminaries}
\noindent\textbf{Counterfactual VQA:} the use of counterfactual inference to mitigate language bias in vision-language tasks dates back to Unbiased SGG~\citep{tang2020unbiased} and CF-VQA~\citep{niu2021counterfactual}. These works were the first to introduce the concepts of Total Direct Effect (TDE) and Total Indirect Effect (TIE) from the field of causality to achieve unbiased estimations via logit subtraction.

Since an LVLM can be regarded as a general VQA model, we take CF-VQA as an example. The TIE-based counterfactual logits can be formulated as:
\begin{equation}
    TIE = Z(q,v,k) - Z(q,v^*,k^*),
\label{eq1-2}
\end{equation}
where $Z(\cdot)$ denotes the model producing answer logits, $q$ denotes the question feature, $v$ is the visual feature, $k$ is the multi-modal fusion feature, $v^*$ and $k^*$ are counterfactual dummy features agnostic to the inputs. In conventional VQA, which is formulated as a closed-set classification task, the unbiased answer is obtained by returning the candidate answer with the highest TIE logits.

\noindent\textbf{Visual Contrastive Decoding (VCD): } building upon the idea of Contrastive Decoding (CD)~\citep{li2023contrastive}, VCD extends CD to mitigate object hallucination during LVLM inference, which can be formulated as follows:
\begin{align}
    p(y|v,v^*,q) &= softmax((1+\alpha) \, \text{logit} (y|v,q) \notag \\
     &- \alpha \, \text{logit} (y|v^*,q)),
\label{eq2}
\end{align}
where $y$ denotes the generated discrete token, $\alpha$ is a trade-off hyperparameter, $q$ and $v$ represent the input textual and visual tokens, respectively, and $v^*$ corresponds to visual tokens obtained from a noisy image. The previously generated tokens are considered part of $q$ for simplicity. The final VCD answer is therefore iteratively sampled from $p(y|v,v^*,q)$.

\subsection{Self-Critical Inference Framework}
In this paper, we observe that VCD essentially reweights the original logits using TIE logits from CF-VQA. Building on this insight, we propose a Self-Critical Inference (SCI) framework, which enhances model robustness through systematic logit-level reasoning over textual and visual counterfactual samples. The proposed SCI framework not only unifies the formulations of VCD and CF-VQA, but also provides a principled solution to both language bias and sensitivity.

We begin by revisiting VCD through the lens of CF-VQA. Specifically, we treat object hallucination in LVLMs as the consequence of iterative biased token generation and frame the decoding process as a sequence of biased classifications. This perspective highlights that LVLMs are fundamentally no different from conventional VQA models. At each generation step, the bias can be mitigated through reasoning over counterfactual logits. Based on this observation, we transform the probability expression in \eqref{eq2} into a logit-based formulation $Z_{vcd}(v,v^*,q)$ as follows: 
\begin{equation}
    Z_{vcd}(v,v^*,q) = (1+\alpha) \, Z(v,q) - \alpha \, Z(v^*,q),
    \label{eq3}
\end{equation}
where $Z(\cdot)$ denotes the LVLM that takes both textual tokens $q$ and visual tokens (either $v$ from real images or $v^*$ from dummy ones) as input and output the logits for the next token. Since there are no explicit multi-modal fusion features in the LVLM inputs, we removes $k$ or $k^*$ in original TIE.

To better understand the relationship between VCD and TIE, we transform the above VCD logits into the $ \exp(\cdot)$ domain. By explicitly expanding the $softmax$ function $exp(x_i) / (\sum_j exp(x_j))$ and omitting the normalization term, we approximate the probability using $p(y) \propto exp(\cdot)$. With this simplification, the VCD probability $p(y|v,v^*,q)$ in \eqref{eq2} can be rewritten as:
\begin{align}
    p(y|v,v^*,q) & = softmax(Z_{vcd}(v,v^*,q)) \notag \\
    & \propto exp(Z(v,q) + \alpha \, ( Z(v,q) - Z(v^*,q) )) \notag \\
    & = exp(Z(v,q)) \cdot exp(\alpha \, ( Z(v,q) - Z(v^*,q) )) \notag \\
    & = exp(Z(v,q)) \cdot exp(TIE / \tau).
    \label{eq5}
\end{align}

The above formulation bridges VCD and CF-VQA, showing that VCD essentially performs weighted token generation upon the original output token probability $p(y|v,q)\propto exp(Z(v,q))$, where TIE logits $exp(TIE / \tau)$ serves as a vocabulary-wise reweighting term, thus forcing the model to rely on visual difference. This formulation also clarifies the role of $\alpha$ in VCD. Neither vanilla CD nor TIE itself requires this additional parameter, because the logit difference itself captures the useful effect of real $v$ over dummy $v^*$. Yet, as a reweighting term, it requires a temperature scaling factor to adjust the trade-off strength, so we further denote $\tau = 1/\alpha$.


To establish a more general robust inference framework, it is also necessary to address the overlooked language sensitivity issue as well. Therefore, we propose SCI framework to  incorporates both a Visual Counterfactual (VC) component, which enhances visual cues similar to TIE, and a Textual Counterfactual (TC) component, which ensures prompt-consistent logits, as follows:
\begin{align}
     & p_{\text{SCI}}(y|\bm{v},\bm{q})  \propto exp(\text{TC} / \tau_1) \cdot exp(\text{VC} / \tau_2), \\
     & \text{TC}_k = max_i(Z_k(v^0,q^i)), \, i=0,1,2,...,N \\
     & \text{VC} = Z(v^0,q^0) - \mathbb{E}[Z(v^j,q^0)], \, j=1,2,...,M 
\end{align}
where $\bm{v}=\{v_j\}_{j=0}^M$ and $\bm{q}=\{q_i\}_{i=0}^N$ denote overall inputs; $M$ and $N$ are the number of visual and textual counterfactual variations, respectively; $v^0$ and $q^0$ stand for original visual and textual tokens; $\{v^j, j\neq 0\}$ and $\{q^i, i\neq 0\}$ represent counterfactual visual tokens generated from content-removed images and counterfactual textual tokens from semantically equivalent but lexically different prompts, respectively. The detailed implementation of these counterfactual samples will be explained in Experiments and Appendix. The operator $max_i(Z_k(\cdot))$ computes the element-wise maximum over $N+1$ samples on the $k-$th dimension of the logits for better consistency. VC enhances the original TIE by incorporating multiple counterfactual visual inputs to obtain a more stable estimation. $\tau_1$ and $\tau_2$ are temperature scaling factor for TC and VC logits, respectively. Following VCD, we also adopt Adaptive Plausibility Constraints as a post process before sampling from $p_{\text{SCI}}(y)$, details will be given in Appendix.

The overall SCI framework provides a generalized solution for robust LVLM inference. In this unified framework, prior works such as VCD and CF-VQA can be viewed as special cases. For VCD, there are no counterfactual prompt variations ($N=0$) and only one counterfactual image ($M=1$). For CF-VQA, the entire TC component is set to a constant and $M=1$. As demonstrated in our experiments, increasing the number of counterfactual inference rounds, \textit{i.e.} using larger $M$ and $N$, leads to more robust final outputs, revealing a new potential test-time scaling strategy for robustness in LVLMs. We also believe that there remains a large opportunity to improve the effectiveness by developing more advanced TC and VC algorithms in future work.


\begin{table}
    \setlength{\tabcolsep}{3pt}
    \centering
    \fontsize{8}{10}\selectfont
    \begin{tabular}{ccccc} 
    \toprule
    Subset Size & B Subset & S Subset & BS Subset & Overlap  \\
    \midrule
    LLaVA-NeXT (MCQ)      & 1810 & 1005 & 2476 & 339 \\
    LLaVA-NeXT (Others)   & 345  & 582  & 794  & 133 \\
    LLaVA-NeXT (Overall)  & 2155 & 1587 & 3270 & 472 \\
    \midrule
    Qwen2-VL (MCQ)        & 1080 & 252 & 1243 & 89 \\
    Qwen2-VL (Others)     & 327  & 311 & 513  & 125 \\
    Qwen2-VL (Overall)    & 1407 & 563 & 1756 & 214 \\
    \bottomrule
    \end{tabular}
    \caption{The size of each subset in constructed DRBench. The overall number of samples across all 6 datasets is 13251, with MCQ and Others categories being 10632 and 2619, respectively.}
    \label{tab-data}
\end{table}




\subsection{Dynamic Robustness Benchmark}
Collecting and constructing datasets tailored to specific robust issues is often cumbersome and costly. What's worse, once such datasets are publicly released, they may be inadvertently integrated into the web-crawled training corpus of subsequent LVLMs. To better evaluate language bias and sensitivity in real downstream tasks, we introduce the Dynamic Robustness Benchmark (DRBench), guided by two main motivations: 1) the evaluation benchmark should be model-specific and dynamic. Since different LVLMs may exhibit varying levels of robustness and their vulnerable samples are not the same, it is important to disentangle the confounding effect of the base model performance from the improvements brought by different inference strategies, so we can better understand the contribution of the inference algorithm itself; 2) existing LVLM bias evaluation datasets typically focus on a single question type and adopt formats that differ significantly from real-world LVLM tasks, \textit{e.g.} exist-or-not questions~\citep{li2023hallucination}. Therefore, it is necessary to develop methods that can automatically adapt to diverse question types and task formats.

Following the above two guiding principles, the proposed benchmark enables the transformation of any popular or newly released LVLM dataset regardless of its question formats into a robustness evaluation benchmark. Specifically, it will adaptively generate model-specific bias subset, sensitivity subset and their union BS Subset for any given LVLM dataset through a two-step process. First, we will evaluate the dataset using the given model. Then, we will adopt the following criteria for filtering the Bias Subset (BS) and the Sensitivity Subset (SS): $\text{BS} = \{(a_{gt},v^0,q^0) \, | \, \forall j\neq 0, \arg\max_a p(a|v^0,q^0) = \arg\max_a p(a|v^j,q^0) \neq a_{gt} \}$ and $\text{SS} = \{(a_{gt},v^0,q^0) \, | \, \forall i\neq 0, \arg\max_a p(a|v^0,q^0) \neq \arg\max_a p(a|v^0,q^i) \}$,
where $a$ and $a_{gt}$ denote the predicted answer and the ground-truth answer, respectively, and $\arg\max_a p(a|\cdot)$ means the predicted answer is obtained via greedy decoding. The generation of counterfactual inputs $v^{j}$ and $q^{i}$ follows the same procedure as in SCI.  In this paper, we fix $M=N=2$ for all our subsets construction. In essence, for BS, we select samples that yield the same incorrect predictions under both the original and dummy visual inputs, indicating a reliance on spurious language priors; for SS, we identify samples whose predictions change in response to subtle, non-causal prompt variations. The final BS Subset is defined as the union of the above two subsets, enabling the investigation of both bias and sensitivity issues. We further split all samples into two groups based on their question types: MCQ for the dominant Multiple-Choice Question type and Others for Yes/No or general open-ended QA types.

In summary, the proposed DRBench offers three key advantages. First, robustness is a model-specific problem, samples that are biased or sensitive for one model may not be vulnerable for another, more evidence will be provided in Table~\ref{tab-appx-cross-model}. An adaptive and model-specific robustness benchmark can thus prevent newly developed LVLMs from being exposed to publicly released fixed datasets and misleading the evaluation of their real underline robustness. Second, as shown in Figure~\ref{fig:tf}(c), different models exhibit varying levels of robustness, the size of each subset provides valuable insight into different models. For example, Table~\ref{tab-data} indicates that: 1) Qwen2-VL is generally more robust than LLaVA-Next; 2) Qwen2-VL is more vulnerable to bias than to sensitivity; and 3) LLaVA-NeXT exhibits more sensitivity issues compared to Qwen2-VL. Third, DRBench enables the evaluation of robustness in various real-world tasks, rather than predefined questions such as a simple exist-or-not (Yes/No) assessment commonly used in previous work~\citep{li2023hallucination}. It also allows for the effortless conversion of any real-world LVLM dataset into the DRBench format, eliminating the need for labor-intensive sample collection and manual annotation.

\section{Experiments}
\label{sec.4}

\subsection{Benchmark Settings}
In our experiments, we construct DRBench using 6 widely adopted LVLM benchmarks: MME~\citep{fu2023mme}, MMStar~\citep{chen2024we}, CCBench~\citep{liu2024mmbench}, ViLP~\citep{luo2024probing}, MMBench-DEV-EN-V11 and MMBench-DEV-CN-V11~\citep{liu2024mmbench}. We begin by randomly splitting the datasets into 20\% validation and 80\% test sets, resulting in 3,315 and 13,251 samples, respectively. Detailed subset statistics are provided in Table~\ref{tab-data}. Note that the size of DRBench increases with larger number of $M$ and $N$. For consistency and convenience, we fix $M=N=2$ for all subsets constructions throughout our experiments. As we mentioned, to enable a more fine-grained analysis, we separately report performance for Multiple-Choice Question (MCQ) and Others (Open-ended QA for ViLP or Yes/No for MME) categories, in addition to the overall results. We use top-1 accuracy as the evaluation metric for all experiments. For the MME dataset, which adopts a different scoring metric, we convert its results to accuracy, so they can be integrated with samples from other datasets to get the final results.


\subsection{Implementation Details}
\noindent\textbf{Model Zoo.} We used Hugging Face versions of Qwen2-VL-7B-Instruct~\citep{Qwen2-VL} and Llama3-LLaVa-NeXT-8B-hf~\citep{liu2024llavanext} as our base models. Following their default configurations, the experiments were conducted using bfloat16 precision and top-k sampling decoding for Qwen2-VL, while LLaVa-NeXT used float16 and greedy decoding. 

\noindent\textbf{Environments.} All experiments were conducted using VLMEvalKit~\citep{duan2024vlmevalkit} on a single NVIDIA A800 GPU of 80GB memory with environment: Pytorch=2.6, Transformers=4.49, and Flash Attention=2.7~\citep{dao2023flashattention}.

\noindent\textbf{Algorithm details.} We evaluated 4 inference strategies: TIE, VCD, M3ID, and the proposed SCI. We adapted Total Indirect Effect (TIE) from CF-VQA~\citep{niu2021counterfactual} to LVLMs by removing the multi-modal features $k$ and $k^*$ in Eq.~\eqref{eq1-2}. For fair comparison, we also incorporated the Adaptive Plausibility Constraints used in VCD and M3ID into TIE. VCD~\citep{leng2024mitigating} and M3ID~\citep{favero2024m3id} share the same mathematical formulation as Eq.~\eqref{eq5}, except that the hyperparameter $\tau$ in M3ID varies depending on the position of the predicted token. For the proposed SCI, we added subscripts such as SCI$_3$, SCI$_5$, and SCI$_7$ to indicate the number of inference rounds. For example, SCI$_5$ means that the total number of counterfactual visual and textual variations, together with the original inputs is 5, \textit{i.e.}, $M+N+1=5$ In our experiments, we set $M=N=1$, $M=N=2$, and $M=N=3$ for SCI$_3$, SCI$_5$, and SCI$_7$, respectively. 

\begin{table*}
    \centering
    \begin{tabular}{lccccccccccc} 
    \toprule
    \multirow{2}{*}{\textbf{Method}}   
          & \multicolumn{3}{c}{\textbf{B Subset}}  
          &  & \multicolumn{3}{c}{\textbf{S Subset}} 
          &  & \multicolumn{3}{c}{\textbf{BS Subset}}  \\ 
    \cmidrule{2-4}\cmidrule{6-8}\cmidrule{10-12}
    & MCQ & Others & Overall & & MCQ & Others & Overall & & MCQ & Others & Overall \\
    \midrule
    LLaVA-NeXT & 0.0 & 0.0 & 0.0 & & 39.2 & 37.63 & 38.63 & & 15.91 & 27.58 & 18.75         \\
    LLaVA-NeXT-TIE & 12.98 & 23.48 & 14.66 & & 39.00 & 57.56 & 45.81 & & 21.89 & 44.21 & 27.31    \\
    LLaVA-NeXT-VCD & 12.65 & 25.51 & 14.71 & & \underline{40.50} & 56.53 & 46.38 & & 22.54 & 44.58 & 27.89   \\
    LLaVA-NeXT-M3ID & 16.91 & 25.22 & 18.24 & & 39.90 & 56.36 & 45.94 & & 24.15 & 44.33 & 29.05  \\
    \cellcolor{lightCyan}LLaVA-NeXT-SCI$_3$ (ours) & \cellcolor{lightCyan}21.22 & \cellcolor{lightCyan}35.36 & \cellcolor{lightCyan}23.48 & \cellcolor{lightCyan} & \cellcolor{lightCyan}39.60 & \cellcolor{lightCyan}\underline{60.31} &  \cellcolor{lightCyan}47.20 & \cellcolor{lightCyan} & \cellcolor{lightCyan}27.14 & \cellcolor{lightCyan}50.13 & \cellcolor{lightCyan}32.72 \\
    
    \cellcolor{lightCyan}LLaVA-NeXT-SCI$_5$ (ours) & \cellcolor{lightCyan}\underline{23.81} & \cellcolor{lightCyan}\underline{37.97} & \cellcolor{lightCyan}\underline{26.08} & \cellcolor{lightCyan} & \cellcolor{lightCyan}\textbf{40.60} & \cellcolor{lightCyan}\textbf{60.65} & \cellcolor{lightCyan}\textbf{47.95} & \cellcolor{lightCyan} & \cellcolor{lightCyan}\underline{28.80} & \cellcolor{lightCyan}\underline{51.01} & \cellcolor{lightCyan}\underline{34.19} \\
    
    \cellcolor{lightCyan}LLaVA-NeXT-SCI$_7$ (ours) & \cellcolor{lightCyan}\textbf{24.86} & \cellcolor{lightCyan}\textbf{38.26} & \cellcolor{lightCyan}\textbf{27.01} & \cellcolor{lightCyan} & \cellcolor{lightCyan}40.10 & \cellcolor{lightCyan}\textbf{60.65} & \cellcolor{lightCyan}\underline{47.64} & \cellcolor{lightCyan} & \cellcolor{lightCyan}\textbf{29.68} & \cellcolor{lightCyan}\textbf{51.26} & \cellcolor{lightCyan}\textbf{34.92}\\
    
    \midrule
    
    Qwen2-VL   & 5.37 & 8.56 & 6.11 & & 38.10 & 34.41 & 36.06 & & 10.78 & 23.59 & 14.52    \\ 
    Qwen2-VL-TIE & 16.20 & 16.82 & 16.35 & & 45.63 & 36.66 & 40.67 & & 20.27 & 27.29 & 22.32  \\
    Qwen2-VL-VCD & 15.74 & 21.71 & 17.13 & & \underline{46.83} & 40.84 & 43.52 & & 20.11 & 30.41 & 23.12 \\
    Qwen2-VL-M3ID & 19.81 & 21.71 & 20.26 & & \textbf{47.22} & 41.16 & 43.87 & & 23.65 & 30.6 & 25.68 \\
    \cellcolor{lightCyan}Qwen2-VL-SCI$_3$ (ours) & \cellcolor{lightCyan}21.67 & \cellcolor{lightCyan}\underline{26.30} & \cellcolor{lightCyan}22.74 & \cellcolor{lightCyan} & \cellcolor{lightCyan}44.05 & \cellcolor{lightCyan}\underline{42.44} & \cellcolor{lightCyan}43.16 & \cellcolor{lightCyan} & \cellcolor{lightCyan}24.54 & \cellcolor{lightCyan}32.75 & \cellcolor{lightCyan}26.94 \\
    
    \cellcolor{lightCyan}Qwen2-VL-SCI$_5$ (ours) & \cellcolor{lightCyan}\underline{24.91} & \cellcolor{lightCyan}25.69 & \cellcolor{lightCyan}\underline{25.09} & \cellcolor{lightCyan} & \cellcolor{lightCyan}\textbf{47.22} & \cellcolor{lightCyan}\underline{42.44} & \cellcolor{lightCyan}\underline{44.58} & \cellcolor{lightCyan} & \cellcolor{lightCyan}\underline{28.00} & \cellcolor{lightCyan}\underline{33.14} & \cellcolor{lightCyan}\underline{29.50}\\
    
    \cellcolor{lightCyan}Qwen2-VL-SCI$_7$ (ours) & \cellcolor{lightCyan}\textbf{27.04} & \cellcolor{lightCyan}\textbf{29.66} & \cellcolor{lightCyan}\textbf{27.65} & \cellcolor{lightCyan} & \cellcolor{lightCyan}\textbf{47.22} & \cellcolor{lightCyan}\textbf{45.98} & \cellcolor{lightCyan}\textbf{46.54} & \cellcolor{lightCyan} & \cellcolor{lightCyan}\textbf{29.61} & \cellcolor{lightCyan}\textbf{36.84} & \cellcolor{lightCyan}\textbf{31.72}\\
    \bottomrule
    \end{tabular}
    \caption{Experiments on B(ias) Subset, S(ensitivity) Subset, and BS Subset of the proposed DRBench. \textbf{Bold texts} indicate the best result of each column and \underline{underline texts} indicate the second best result.}
    \label{tab-main}
\end{table*}

\noindent\textbf{Counterfactual sample construction.} We constructed up to 3 visual counterfactual variations and 3 prompt variations: 1) VC-Color0(C0) renders the input image into black; 2) VC-Noise500(N500) and 3) VC-Noise400 apply the diffusion noise function from VCD, using noise steps of 500 and 400, respectively; 3) TC-V1 adds an additional system prompt instructing the model to focus on image details; 4) TC-V2 further modifies the system prompt’s language from English to Chinese or vice versa; 5) TC-V3 that injects identity information by prompting the model to respond as a clever student. More detailed prompts will be given in the Appendix.

\begin{table}
    \centering
    \fontsize{7}{10}\selectfont
    \begin{tabular}{llccc} 
    \toprule
    Construction Model & Methods & MCQ & Others & Overall \\
    \midrule
    \multirow{4}{*}{LLaVA-NeXT} & LLaVA-NeXT-Original & 15.91 & 27.58 & 18.75 \\
    & LLaVA-NeXT-SCI$_5$ & 28.80 & 51.01 & 34.19 \\
    & Qwen2-VL-Original & 59.29 & 63.48 & 60.31 \\ 
    & Qwen2-VL-SCI$_5$ & 61.15 & 67.88 & 62.78 \\
    
    \midrule
    
    \multirow{4}{*}{Qwen2-VL} & Qwen2-VL-Original & 10.78 & 23.59 & 14.52 \\ 
    & Qwen2-VL-SCI$_5$ & 28.00 & 33.14 & 29.50 \\
    & LLaVA-NeXT-Original & 30.25 & 39.18 & 32.86 \\
    & LLaVA-NeXT-SCI$_5$ & 34.59 & 41.33 & 36.56 \\
    \bottomrule
    \end{tabular}
    \caption{Ablation on cross-model BS Subset evaluation.}
    \label{tab-appx-cross-model}
\end{table}

\noindent\textbf{Hyperparameter settings.} Based on the validation results, we set $\tau_1$ to $1.5$, $2$, and $2.5$ for SCI$_3$, SCI$_5$, and SCI$_7$, respectively. Since the TC component involves element-wise maximum over logits, its magnitude increases with the number of variations $N$. Therefore, the temperature scaling factor $\tau_1$ should be increased accordingly to maintain a similar distribution of TC logits. The $\tau_2$ is fixed at $0.2$, because the averaging operation in the VC logits stabilizes the distribution and mitigates the need for the scaling change. For the Adaptive Plausibility Constraint~\citep{leng2024mitigating} used in our experiments, the threshold parameter is set to $0.3$ unless otherwise specified. More details about the constraint and hyperparameter ablation will be introduced in the Appendix.

\subsection{Experimental Results}
\noindent\textbf{Experiments on the proposed DRBench.} As shown in Table~\ref{tab-main}, we adopted two state-of-the-art LVLMs for our experiments: LLaVA-NeXT-8B and Qwen2-VL-7B. We compared the base model performances and three other algorithms: TIE, VCD, and M3ID that utilized counterfactual inference. The proposed methods, SCI$_3$, SCI$_5$, and SCI$_7$, consistently demonstrated superior performance across the B(ias), S(ensitivity), and combined BS Subsets. We further reported MCQ and Others results based on question types and saw that the improvements brought by SCI were consistent across both categories. Table~\ref{tab-main} also reveals that the proposed DRBench can successfully disentangle the base model performance and focus on investigating the effectiveness of inference algorithms, as Qwen2-VL outperforms LLaVA-NeXT by 10.19\% on the original datasets in Table~\ref{tab-orig}, while their base and final overall performances on the DRBench are very close.

\begin{table*}
    \centering
    \begin{tabular}{lcccccccccc} 
    \toprule
    \multirow{2}{*}{\textbf{Method}} 
    & \multicolumn{6}{c}{\textbf{Single Dataset}} 
    & & \multicolumn{3}{c}{\textbf{Question Type}} \\
    \cmidrule{2-7}\cmidrule{9-11}
     & MMB-C & MMB-E & MME & CCB & MMS & ViLP & & \textbf{MCQ} & \textbf{Others} & \textbf{Overall} \\ 
    \midrule
    LLaVA-NeXT & 78.0 &	79.72 & 79.57 & 47.0 & 44.75 & 51.53 & & 70.12 & 71.86 & 70.46 \\
    LLaVA-NeXT-TIE & {\color{blue}78.28} & {\color{blue}80.28} & 77.30 & 45.65 & {\color{blue}46.00} & {\color{blue}53.19} & & {\color{blue}70.36} & 70.68 & 70.42 \\
    LLaVA-NeXT-VCD & {\color{blue}78.38} & {\color{blue}80.28} & 78.09 & 46.63 & {\color{blue}45.00} & {\color{blue}54.31} & & {\color{blue}70.44} & 71.55 & {\color{blue}70.66} \\
    LLaVA-NeXT-M3ID & {\color{blue}78.31} & {\color{blue}80.18} & 78.62 & 45.89 & {\color{blue}45.92} & {\color{blue}54.03} & & {\color{blue}70.36} & 71.86 & {\color{blue}70.66} \\
    \cellcolor{lightCyan}LLaVA-NeXT-SCI$_5$ (ours) & \cellcolor{lightCyan}{\color{blue}78.21} & \cellcolor{lightCyan}{\color{blue}80.08} & \cellcolor{lightCyan}{\color{blue}80.15} & \cellcolor{lightCyan}46.20 & \cellcolor{lightCyan}{\color{blue}45.75} & \cellcolor{lightCyan}{\color{blue}53.06} & \cellcolor{lightCyan} & \cellcolor{lightCyan}{\color{blue}70.32} & \cellcolor{lightCyan}{\color{blue}72.70} & \cellcolor{lightCyan}{\color{blue}70.79} \\
    \midrule
    Qwen2-VL & 85.26 & 86.36 & 87.89 & 73.22 & 59.50 & 56.53 & & 80.91 & 79.27 & 80.58\\ 
    Qwen2-VL-TIE & {\color{blue}86.00} & {\color{blue}86.59} & 86.52 & {\color{blue}73.84} & 59.00 & {\color{blue}57.08} & & {\color{blue}81.30} & 78.43 & {\color{blue}80.73}\\
    Qwen2-VL-VCD & {\color{blue}86.05} & {\color{blue}86.56} & 86.41 & {\color{blue}73.77} & {\color{blue}60.08} & {\color{blue}57.92} & & {\color{blue}81.42} & 78.58 & {\color{blue}80.86}\\ 
    Qwen2-VL-M3ID & {\color{blue}85.69} & {\color{blue}86.46} & 86.10 & {\color{blue}73.96} & {\color{blue}59.75} & {\color{blue}57.78} & & {\color{blue}81.25} & 78.31 & {\color{blue}80.67}\\
    \cellcolor{lightCyan}Qwen2-VL-SCI$_5$ (ours) & \cellcolor{lightCyan}{\color{blue}85.97} & \cellcolor{lightCyan}{\color{blue}86.67} & \cellcolor{lightCyan}87.36 & \cellcolor{lightCyan}{\color{blue}73.59} & \cellcolor{lightCyan}{\color{blue}59.92} & \cellcolor{lightCyan}{\color{blue}58.06} & \cellcolor{lightCyan} & \cellcolor{lightCyan}{\color{blue}81.39} & \cellcolor{lightCyan}{\color{blue}79.31} & \cellcolor{lightCyan}{\color{blue}80.98} \\
    \bottomrule
    \end{tabular}
    \caption{Experiments on MMB(ench-Dev)-C/E(N-V11), MME, CCB(ench), MMS(tar), and ViLP indicate that \textbf{SCI has more consistent improvement} than TDE/VCD/M3ID on those real-world LVLM benchmarks (using 80\% test splits). {\color{blue}Blue texts} indicate an improvement over the baseline.}
    \label{tab-orig}
\end{table*}

\noindent\textbf{Experiments on real-world LVLM datasets.} We further evaluated the proposed SCI on 6 popular LVLM datasets to verify its performance under real-world data distributions, in addition to the proposed subsets alone. Taking SCI$_5$ as an example in Table~\ref{tab-orig}, it consistently outperformed the baseline models in all question types and almost all datasets. Meanwhile, TIE, VCD and M3ID decrease the performance on Others question type. Note that although the improvements appear relatively marginal, since vulnerable samples comprise only a portion of the datasets. These results confirm that the gains observed on DRBench are not due to overfitting to specific data distributions, but rather reflect a general improvement in robustness.

\begin{table}
    \setlength{\tabcolsep}{3pt}
    \centering
    \fontsize{9}{10}\selectfont
    \begin{tabular}{ccccccccc} 
    \toprule
    Base & VC-C & VC-N & TC-V1 & TC-V2 & MCQ & Others & Overall \\
    \midrule
     \checkmark & & & & & 10.78 & 23.59 & 14.52 \\
     & \checkmark & & & & 8.77 & 18.52 & 11.62 \\ 
     & & \checkmark & & & 10.62 & 25.15 & 14.86 \\
     & & & \checkmark & & 10.38 & 24.37 & 14.46 \\
     & & & & \checkmark & 12.07 & 23.00 & 15.26 \\
     \checkmark & \checkmark & & & & 21.72 & 29.43 & 23.97 \\
     \checkmark & & & \checkmark & & 10.54 & 23.39 & 14.29 \\
     \checkmark & \checkmark & & \checkmark & & 24.54 & 32.75 & 26.94 \\
     \checkmark & & \checkmark & & \checkmark & 26.67 & 30.97 & 27.93 \\
     \checkmark & \checkmark & \checkmark & & & 27.37 & 31.13 & 28.46 \\
     \checkmark & & & \checkmark & \checkmark & 11.58 & 23.21 & 14.98 \\
     \checkmark & \checkmark & & \checkmark & \checkmark & 26.07 & 32.55 & 27.96 \\
     \checkmark & \checkmark & \checkmark & \checkmark &  & 26.71 & 33.33 & 28.64 \\
     \checkmark & \checkmark & \checkmark & \checkmark & \checkmark & 28.00 & 33.14 & 29.50 \\
    \bottomrule
    \end{tabular}
    \caption{Ablation experiments for different counterfactual logits combinations using Qwen2-VL on BS Subset.}
    \label{tab-ablation-a}
\end{table}

\begin{figure*}[t]
    \centering
    \includegraphics[width=1.0\linewidth]{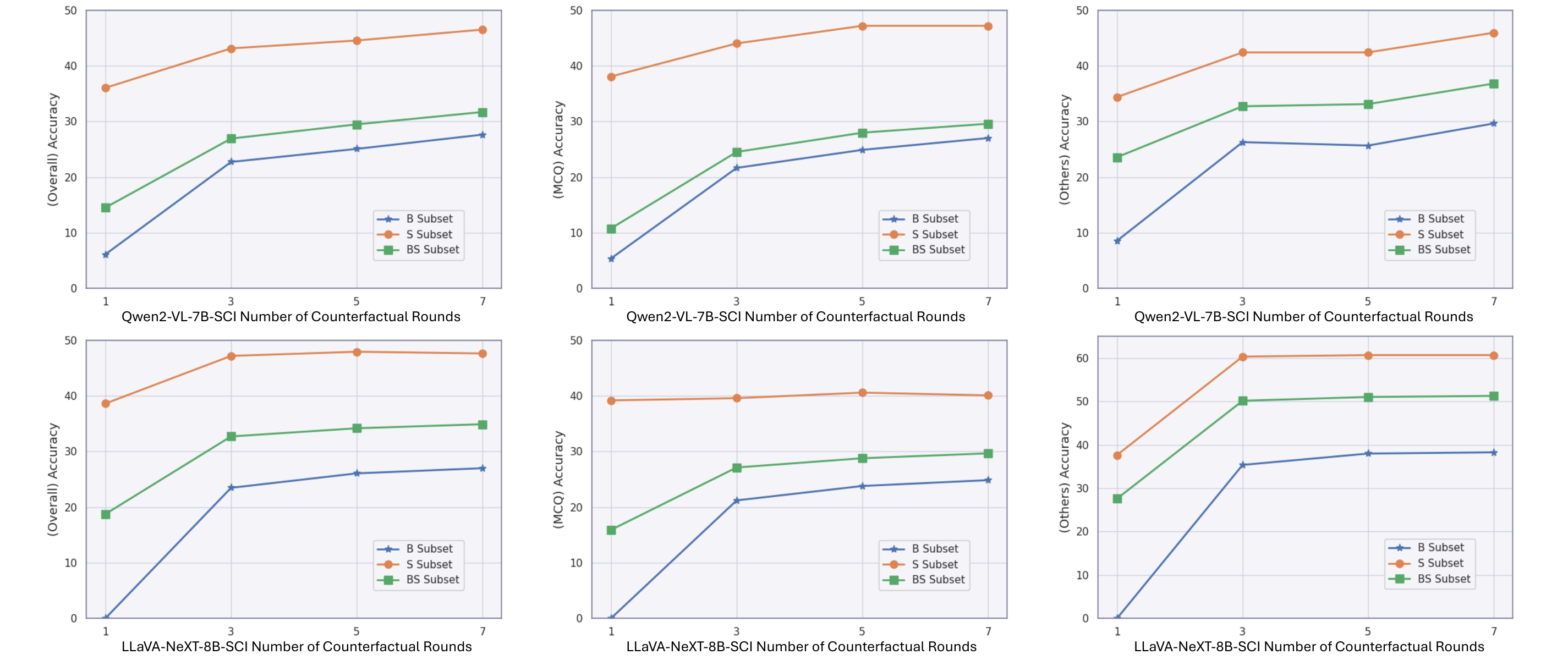}
    \caption{Investigating the test-time scaling effect on robustness with respect to the number of inference rounds on B/S/BS subsets across different question types and LVLMs.}
    \label{fig:2}
\end{figure*}

\noindent\textbf{Ablation study on test-time scaling effect with increasing inference rounds.} To better understand the effect of each component in SCI framework, we conducted an ablation study on SCI$_5$. As shown in Table~\ref{tab-ablation-a}, we first evaluated the performance of the base inputs and four individual counterfactual inputs on the BS Subset. We then incrementally increased the number of counterfactual rounds to form progressively more complete versions of SCI to reach SCI$_5$. Note that experiments on VC component and TC component alone are also included. Together with the comprehensive results of SCI$_3$, SCI$_5$, and SCI$_7$ in Figure~\ref{fig:2}, the overall findings highlight the potential of test-time scaling: robustness of models can be improved with more incorporated counterfactual rounds.

\noindent\textbf{Ablation study on cross-model DRBench evaluation.} The ablation study in Table~\ref{tab-appx-cross-model} provides additional insights: 1) non-robust samples vary significantly across different LVLMs. For instance, the BS Subset constructed by LLaVA-NeXT yields only 18.75\% accuracy on its own model, while Qwen2-VL achieves 60.31\% accuracy on the same subset, and vice versa. This demonstrates that even if an LVLM performs perfectly well on a fixed robustness benchmark, it may still fail on new vulnerable samples. These findings highlight the necessity of adopting a model-specific DRBench; 2) The performance gains achieved through SCI in one model are transferable to DRBenchs constructed by other models, thereby validating the generalization ability of SCI. 

\subsection{Discussions}
We also provide some interesting discussions to shed lights on the proposed SCI framework and DRBench.

\noindent\textbf{Q1: Why did the base models perform so poorly (\textit{e.g.}, LLaVA-NeXT even got 0.0 on the Bias Subset) on the Bias, Sensitivity, and BS Subsets? }

\noindent\textbf{A1:} The proposed DRBench are intentionally designed to probe samples particularly vulnerable to robust issues, \textit{i.e.} they are hard examples for LVLMs. That's why the model performances on these subsets are sometimes even lower than random guessing, \textit{e.g.}, MCQs have a 25\% chance accuracy for random guess. In fact, according to the definition, the Bias Subset specifically collects samples for which the base model consistently produces incorrect predictions, so its accuracy is theoretically expected to be 0.0. The reason why Qwen2-VL does not yield exactly 0.0 is due to its use of top-k sampling for decoding by default. In contrast, LLaVA-NeXT uses greedy decoding, producing deterministic predictions, which explains its consistent 0.0 accuracy on the Bias Subset.

\noindent\textbf{Q2: What's the computational overhead of SCI and are there potential solutions for acceleration? }

\noindent\textbf{A2:} All test-time scaling strategies entail a trade-off between inference time and performance, which means that the proposed SCI will inevitably take more time. The most intuitive acceleration method for SCI is batch inference. Based on our experiments, the computational overhead of SCI$_3$, SCI$_5$, and SCI$_7$ using batch inference is approximately $1.29 \times$, $1.81 \times$, and $2.48 \times$ that of the base model, respectively, which is much faster than the vanilla version, which costs $2.96 \times$, $5.01 \times$, and $6.68 \times$, respectively. We also believe that KV Cache sharing for the visual and textual tokens that remain unchanged is a potential acceleration technique for SCI.


\noindent\textbf{Q3: Why is SCI different from previous test-time scaling studies, and could it open up a new paradigm? }

\noindent\textbf{A3:} Most of the existing test-time scaling studies~\citep{snell2025scaling} focus on increasing the length of intermediate thinking tokens. However, the prompt-level improvement only reveals whether the answer is correct or wrong. By introducing SCI, we go beyond discrete token outputs and analyze the underlying continuous logit distributions through comparison and aggregation of counterfactual logits. This approach provides significantly richer information than simply using final predicted tokens. Therefore, we believe that SCI opens up a promising new direction for test-time scaling studies.


\noindent\textbf{Q4: Does the performance gains of the proposed SCI come from hacking the corresponding DRBench?} 

\noindent\textbf{A4:} Given the overlap process between the DRBench construction and the counterfactual sample construction of SCI, it is possible that the performance gains of SCI are tailored to its corresponding DRBench data. However, results in Table~\ref{tab-appx-cross-model} demonstrate that SCI still yields consistent improvements when evaluated on the vulnerable test sets derived from other models, even though the relative improvements become smaller. This suggests that the proposed SCI possesses inherent generalization capability beyond its corresponding model-specific DRBench.


\section{Conclusion}


In this paper, we propose SCI, a generalized framework for robust inference in LVLMs that jointly addresses language bias and sensitivity through comprehensive logit-level counterfactual reasoning. Complemented by the DRBench, our contributions offer both a methodological advancement and an adaptive evaluation protocol for improving LVLM robustness. Extensive experiments further reveal a scalable pathway toward enhanced test-time robustness, which could be achieved by incorporating more counterfactual inference rounds and advanced logit-level reasoning algorithms. We hope that SCI and DRBench will serve as foundational paradigms and diagnostic tools for the development of future more reliable and trustworthy LVLMs.

\section*{Acknowledgements}
This work was supported by the Double First-Class Initiative Fund, Disciplinary Development Program of the Institute of AI for Engineering, Tongji University.

\clearpage

{
    \small
    \bibliographystyle{ieeenat_fullname}
    \bibliography{egbib}

@String(CVPR= {IEEE Conf. Comput. Vis. Pattern Recog.})

@String(ICLR = {Int. Conf. Learn. Represent.})

@String(AAAI = {AAAI})

@String(CVPR  = {CVPR})

@String(ICLR  = {ICLR})

@article{brown2020language,
  title={Language models are few-shot learners},
  author={Brown, Tom and Mann, Benjamin and Ryder, Nick and Subbiah, Melanie and Kaplan, Jared D and Dhariwal, Prafulla and Neelakantan, Arvind and Shyam, Pranav and Sastry, Girish and Askell, Amanda and others},
  journal={Advances in neural information processing systems},
  volume={33},
  pages={1877--1901},
  year={2020}
}

@article{achiam2023gpt,
  title={Gpt-4 technical report},
  author={Achiam, Josh and Adler, Steven and Agarwal, Sandhini and Ahmad, Lama and Akkaya, Ilge and Aleman, Florencia Leoni and Almeida, Diogo and Altenschmidt, Janko and Altman, Sam and Anadkat, Shyamal and others},
  journal={arXiv preprint arXiv:2303.08774},
  year={2023}
}

@article{touvron2023llama,
  title={Llama: Open and efficient foundation language models},
  author={Touvron, Hugo and Lavril, Thibaut and Izacard, Gautier and Martinet, Xavier and Lachaux, Marie-Anne and Lacroix, Timoth{\'e}e and Rozi{\`e}re, Baptiste and Goyal, Naman and Hambro, Eric and Azhar, Faisal and others},
  journal={arXiv preprint arXiv:2302.13971},
  year={2023}
}

@article{bai2023qwen,
  title={Qwen technical report},
  author={Bai, Jinze and Bai, Shuai and Chu, Yunfei and Cui, Zeyu and Dang, Kai and Deng, Xiaodong and Fan, Yang and Ge, Wenbin and Han, Yu and Huang, Fei and others},
  journal={arXiv preprint arXiv:2309.16609},
  year={2023}
}

@article{liu2024deepseek,
  title={Deepseek-v3 technical report},
  author={Liu, Aixin and Feng, Bei and Xue, Bing and Wang, Bingxuan and Wu, Bochao and Lu, Chengda and Zhao, Chenggang and Deng, Chengqi and Zhang, Chenyu and Ruan, Chong and others},
  journal={arXiv preprint arXiv:2412.19437},
  year={2024}
}

@misc{liu2024llavanext,
    title={LLaVA-NeXT: Improved reasoning, OCR, and world knowledge},
    url={https://llava-vl.github.io/blog/2024-01-30-llava-next/},
    author={Liu, Haotian and Li, Chunyuan and Li, Yuheng and Li, Bo and Zhang, Yuanhan and Shen, Sheng and Lee, Yong Jae},
    month={January},
    year={2024}
}

@misc{liu2023llava,
      title={Visual Instruction Tuning}, 
      author={Liu, Haotian and Li, Chunyuan and Wu, Qingyang and Lee, Yong Jae},
      publisher={NeurIPS},
      year={2023},
}

@article{Qwen-VL,
  title={Qwen-VL: A Versatile Vision-Language Model for Understanding, Localization, Text Reading, and Beyond},
  author={Bai, Jinze and Bai, Shuai and Yang, Shusheng and Wang, Shijie and Tan, Sinan and Wang, Peng and Lin, Junyang and Zhou, Chang and Zhou, Jingren},
  journal={arXiv preprint arXiv:2308.12966},
  year={2023}
}

@article{Qwen2-VL,
  title={Qwen2-VL: Enhancing Vision-Language Model's Perception of the World at Any Resolution},
  author={Wang, Peng and Bai, Shuai and Tan, Sinan and Wang, Shijie and Fan, Zhihao and Bai, Jinze and Chen, Keqin and Liu, Xuejing and Wang, Jialin and Ge, Wenbin and Fan, Yang and Dang, Kai and Du, Mengfei and Ren, Xuancheng and Men, Rui and Liu, Dayiheng and Zhou, Chang and Zhou, Jingren and Lin, Junyang},
  journal={arXiv preprint arXiv:2409.12191},
  year={2024}
}

@article{yin2024survey,
  title={A survey on multimodal large language models},
  author={Yin, Shukang and Fu, Chaoyou and Zhao, Sirui and Li, Ke and Sun, Xing and Xu, Tong and Chen, Enhong},
  journal={National Science Review},
  volume={11},
  number={12},
  year={2024}
}

@inproceedings{tang2020unbiased,
  title={Unbiased scene graph generation from biased training},
  author={Tang, Kaihua and Niu, Yulei and Huang, Jianqiang and Shi, Jiaxin and Zhang, Hanwang},
  booktitle={Proceedings of the IEEE/CVF conference on computer vision and pattern recognition},
  pages={3716--3725},
  year={2020}
}

@article{wang2022self,
  title={Self-consistency improves chain of thought reasoning in language models},
  author={Wang, Xuezhi and Wei, Jason and Schuurmans, Dale and Le, Quoc and Chi, Ed and Narang, Sharan and Chowdhery, Aakanksha and Zhou, Denny},
  journal={arXiv preprint arXiv:2203.11171},
  year={2022}
}

@article{kwon2024stableprompt,
  title={StablePrompt: Automatic Prompt Tuning using Reinforcement Learning for Large Language Models},
  author={Kwon, Minchan and Kim, Gaeun and Kim, Jongsuk and Lee, Haeil and Kim, Junmo},
  journal={arXiv preprint arXiv:2410.07652},
  year={2024}
}

@article{pitis2023boosted,
  title={Boosted prompt ensembles for large language models},
  author={Pitis, Silviu and Zhang, Michael R and Wang, Andrew and Ba, Jimmy},
  journal={arXiv preprint arXiv:2304.05970},
  year={2023}
}

@inproceedings{chen2020counterfactual,
  title={Counterfactual samples synthesizing for robust visual question answering},
  author={Chen, Long and Yan, Xin and Xiao, Jun and Zhang, Hanwang and Pu, Shiliang and Zhuang, Yueting},
  booktitle={Proceedings of the IEEE/CVF conference on computer vision and pattern recognition},
  pages={10800--10809},
  year={2020}
}

@inproceedings{gunjal2024detecting,
  title={Detecting and preventing hallucinations in large vision language models},
  author={Gunjal, Anisha and Yin, Jihan and Bas, Erhan},
  booktitle={Proceedings of the AAAI Conference on Artificial Intelligence},
  volume={38},
  pages={18135--18143},
  year={2024}
}

@inproceedings{goyal2017making,
  title={Making the v in vqa matter: Elevating the role of image understanding in visual question answering},
  author={Goyal, Yash and Khot, Tejas and Summers-Stay, Douglas and Batra, Dhruv and Parikh, Devi},
  booktitle={Proceedings of the IEEE conference on computer vision and pattern recognition},
  pages={6904--6913},
  year={2017}
}

@inproceedings{zhang2024mm,
  title={MM-LLMs: Recent Advances in MultiModal Large Language Models},
  author={Zhang, Duzhen and Yu, Yahan and Dong, Jiahua and Li, Chenxing and Su, Dan and Chu, Chenhui and Yu, Dong},
  booktitle={Findings of the Association for Computational Linguistics ACL 2024},
  pages={12401--12430},
  year={2024}
}

@misc{dai2023instructblp,
      title={InstructBLIP: Towards General-purpose Vision-Language Models with Instruction Tuning}, 
      author={Wenliang Dai and Junnan Li and Dongxu Li and Anthony Meng Huat Tiong and Junqi Zhao and Weisheng Wang and Boyang Li and Pascale Fung and Steven Hoi},
      year={2023},
      eprint={2305.06500},
      archivePrefix={arXiv},
      primaryClass={cs.CV},
      url={https://arxiv.org/abs/2305.06500}, 
}

@inproceedings{antol2015vqa,
  title={Vqa: Visual question answering},
  author={Antol, Stanislaw and Agrawal, Aishwarya and Lu, Jiasen and Mitchell, Margaret and Batra, Dhruv and Zitnick, C Lawrence and Parikh, Devi},
  booktitle={Proceedings of the IEEE international conference on computer vision},
  pages={2425--2433},
  year={2015}
}

@inproceedings{duan2024vlmevalkit,
  title={Vlmevalkit: An open-source toolkit for evaluating large multi-modality models},
  author={Duan, Haodong and Yang, Junming and Qiao, Yuxuan and Fang, Xinyu and Chen, Lin and Liu, Yuan and Dong, Xiaoyi and Zang, Yuhang and Zhang, Pan and Wang, Jiaqi and others},
  booktitle={Proceedings of the 32nd ACM International Conference on Multimedia},
  pages={11198--11201},
  year={2024}
}

@article{dao2023flashattention,
  title={Flashattention-2: Faster attention with better parallelism and work partitioning},
  author={Dao, Tri},
  journal={arXiv preprint arXiv:2307.08691},
  year={2023}
}

@article{luo2024probing,
  title={Probing Visual Language Priors in VLMs},
  author={Luo, Tiange and Cao, Ang and Lee, Gunhee and Johnson, Justin and Lee, Honglak},
  journal={arXiv preprint arXiv:2501.00569},
  year={2024}
}

@inproceedings{jiang2025devils,
  title={Devils in middle layers of large vision-language models: Interpreting, detecting and mitigating object hallucinations via attention lens},
  author={Jiang, Zhangqi and Chen, Junkai and Zhu, Beier and Luo, Tingjin and Shen, Yankun and Yang, Xu},
  booktitle={Proceedings of the Computer Vision and Pattern Recognition Conference},
  pages={25004--25014},
  year={2025}
}

@inproceedings{li2023evaluating,
  title={Evaluating Object Hallucination in Large Vision-Language Models},
  author={Li, Yifan and Du, Yifan and Zhou, Kun and Wang, Jinpeng and Zhao, Xin and Wen, Ji-Rong},
  booktitle={The 2023 Conference on Empirical Methods in Natural Language Processing},
  year={2024}
}

@article{ho2020denoising,
  title={Denoising diffusion probabilistic models},
  author={Ho, Jonathan and Jain, Ajay and Abbeel, Pieter},
  journal={Advances in neural information processing systems},
  volume={33},
  pages={6840--6851},
  year={2020}
}

@inproceedings{leng2024mitigating,
  title={Mitigating object hallucinations in large vision-language models through visual contrastive decoding},
  author={Leng, Sicong and Zhang, Hang and Chen, Guanzheng and Li, Xin and Lu, Shijian and Miao, Chunyan and Bing, Lidong},
  booktitle={Proceedings of the IEEE/CVF Conference on Computer Vision and Pattern Recognition},
  pages={13872--13882},
  year={2024}
}

@misc{favero2024m3id,
      title={Multi-Modal Hallucination Control by Visual Information Grounding}, 
      author={Alessandro Favero and Luca Zancato and Matthew Trager and Siddharth Choudhary and Pramuditha Perera and Alessandro Achille and Ashwin Swaminathan and Stefano Soatto},
      year={2024},
      archivePrefix={CVPR},
}

@inproceedings{li2023contrastive,
    title = "Contrastive Decoding: Open-ended Text Generation as Optimization",
    author = "Li, Xiang Lisa  and
      Holtzman, Ari  and
      Fried, Daniel  and
      Liang, Percy  and
      Eisner, Jason  and
      Hashimoto, Tatsunori  and
      Zettlemoyer, Luke  and
      Lewis, Mike",
    editor = "Rogers, Anna  and
      Boyd-Graber, Jordan  and
      Okazaki, Naoaki",
    booktitle = "Proceedings of the 61st Annual Meeting of the Association for Computational Linguistics (Volume 1: Long Papers)",
    month = jul,
    year = "2023",
    address = "Toronto, Canada",
    publisher = "Association for Computational Linguistics",
    url = "https://aclanthology.org/2023.acl-long.687/",
    doi = "10.18653/v1/2023.acl-long.687",
    pages = "12286--12312",
}

@misc{kaplan2020scaling,
      title={Scaling Laws for Neural Language Models}, 
      author={Jared Kaplan and Sam McCandlish and Tom Henighan and Tom B. Brown and Benjamin Chess and Rewon Child and Scott Gray and Alec Radford and Jeffrey Wu and Dario Amodei},
      year={2020},
      eprint={2001.08361},
      archivePrefix={arXiv},
      primaryClass={cs.LG},
      url={https://arxiv.org/abs/2001.08361}, 
}

@misc{hoffmann2022training,
      title={Training Compute-Optimal Large Language Models}, 
      author={Jordan Hoffmann and Sebastian Borgeaud and Arthur Mensch and Elena Buchatskaya and Trevor Cai and Eliza Rutherford and Diego de Las Casas and Lisa Anne Hendricks and Johannes Welbl and Aidan Clark and Tom Hennigan and Eric Noland and Katie Millican and George van den Driessche and Bogdan Damoc and Aurelia Guy and Simon Osindero and Karen Simonyan and Erich Elsen and Jack W. Rae and Oriol Vinyals and Laurent Sifre},
      year={2022},
      eprint={2203.15556},
      archivePrefix={arXiv},
      primaryClass={cs.CL},
      url={https://arxiv.org/abs/2203.15556}, 
}

@inproceedings{snell2025scaling,
title={Scaling LLM Test-Time Compute Optimally Can be More Effective than Scaling Parameters for Reasoning},
author={Charlie Victor Snell and Jaehoon Lee and Kelvin Xu and Aviral Kumar},
booktitle={The Thirteenth International Conference on Learning Representations},
year={2025},
url={https://openreview.net/forum?id=4FWAwZtd2n}
}

@article{jiang2023calibrating,
  title={Calibrating language models via augmented prompt ensembles},
  author={Jiang, Mingjian and Ruan, Yangjun and Huang, Sicong and Liao, Saifei and Pitis, Silviu and Grosse, Roger Baker and Ba, Jimmy},
  journal={arXiv preprint},
  year={2023}
}

@inproceedings{arora2023ask,
  title={Ask Me Anything: A simple strategy for prompting language models},
  author={Arora, Simran and Narayan, Avanika and Chen, Mayee F and Orr, Laurel J and Guha, Neel and Bhatia, Kush and Chami, Ines and R{\'e}, Christopher},
  booktitle={ICLR},
  year={2023}
}

@inproceedings{zhou2024analyzing,
  title={ANALYZING AND MITIGATING OBJECT HALLUCINATION IN LARGE VISION-LANGUAGE MODELS},
  author={Zhou, Yiyang and Cui, Chenhang and Yoon, Jaehong and Zhang, Linjun and Deng, Zhun and Finn, Chelsea and Bansal, Mohit and Yao, Huaxiu},
  booktitle={12th International Conference on Learning Representations, ICLR 2024},
  year={2024}
}

@article{radford2019gpt2,
  title={Language Models are Unsupervised Multitask Learners},
  author={Radford, Alec and Wu, Jeff and Child, Rewon and Luan, David and Amodei, Dario and Sutskever, Ilya},
  journal={technical report},
  year={2019}
}

@inproceedings{xu2015show,
  title={Show, attend and tell: Neural image caption generation with visual attention},
  author={Xu, Kelvin and Ba, Jimmy and Kiros, Ryan and Cho, Kyunghyun and Courville, Aaron and Salakhudinov, Ruslan and Zemel, Rich and Bengio, Yoshua},
  booktitle={International conference on machine learning},
  pages={2048--2057},
  year={2015},
  organization={PMLR}
}

@inproceedings{niu2021counterfactual,
  title={Counterfactual vqa: A cause-effect look at language bias},
  author={Niu, Yulei and Tang, Kaihua and Zhang, Hanwang and Lu, Zhiwu and Hua, Xian-Sheng and Wen, Ji-Rong},
  booktitle={Proceedings of the IEEE/CVF conference on computer vision and pattern recognition},
  pages={12700--12710},
  year={2021}
}

@inproceedings{yang2019auto,
  title={Auto-encoding scene graphs for image captioning},
  author={Yang, Xu and Tang, Kaihua and Zhang, Hanwang and Cai, Jianfei},
  booktitle={Proceedings of the IEEE/CVF conference on computer vision and pattern recognition},
  pages={10685--10694},
  year={2019}
}

@inproceedings{zhai2023sigmoid,
  title={Sigmoid loss for language image pre-training},
  author={Zhai, Xiaohua and Mustafa, Basil and Kolesnikov, Alexander and Beyer, Lucas},
  booktitle={Proceedings of the IEEE/CVF international conference on computer vision},
  pages={11975--11986},
  year={2023}
}

@inproceedings{openai2021clip,
  title={Learning transferable visual models from natural language supervision},
  author={Radford, Alec and Kim, Jong Wook and Hallacy, Chris and Ramesh, Aditya and Goh, Gabriel and Agarwal, Sandhini and Sastry, Girish and Askell, Amanda and Mishkin, Pamela and Clark, Jack and others},
  booktitle={International conference on machine learning},
  pages={8748--8763},
  year={2021},
  organization={PmLR}
}

@misc{fu2023mme,
  title={MME: A Comprehensive Evaluation Benchmark for Multimodal Large Language Models}, 
  author={Chaoyou Fu and Peixian Chen and Yunhang Shen and Yulei Qin and Mengdan Zhang and Xu Lin and Jinrui Yang and Xiawu Zheng and Ke Li and Xing Sun and Yunsheng Wu and Rongrong Ji},
  year={2023},
  eprint={2306.13394},
  archivePrefix={arXiv},
  primaryClass={cs.CV},
  url={https://arxiv.org/abs/2306.13394}, 
}

@article{chen2024we,
  title={Are We on the Right Way for Evaluating Large Vision-Language Models?},
  author={Chen, Lin and Li, Jinsong and Dong, Xiaoyi and Zhang, Pan and Zang, Yuhang and Chen, Zehui and Duan, Haodong and Wang, Jiaqi and Qiao, Yu and Lin, Dahua and others},
  journal={arXiv preprint arXiv:2403.20330},
  year={2024}
}

@inproceedings{li2023hallucination,
  title={Evaluating Object Hallucination in Large Vision-Language Models},
  author={Yifan, Li and Yifan, Du and Kun, Zhou and Jinpeng, Wang and Wayne Xin, Zhao and Ji-Rong, Wen},
  booktitle={The 2023 Conference on Empirical Methods in Natural Language Processing},
  year={2023},
  url={https://openreview.net/forum?id=xozJw0kZXF}
}

@inproceedings{liu2024mmbench,
  title={Mmbench: Is your multi-modal model an all-around player?},
  author={Liu, Yuan and Duan, Haodong and Zhang, Yuanhan and Li, Bo and Zhang, Songyang and Zhao, Wangbo and Yuan, Yike and Wang, Jiaqi and He, Conghui and Liu, Ziwei and others},
  booktitle={European conference on computer vision},
  pages={216--233},
  year={2024},
  organization={Springer}
}

@article{wen2021debiased,
  title={Debiased visual question answering from feature and sample perspectives},
  author={Wen, Zhiquan and Xu, Guanghui and Tan, Mingkui and Wu, Qingyao and Wu, Qi},
  journal={Advances in Neural Information Processing Systems},
  volume={34},
  pages={3784--3796},
  year={2021}
}

@misc{suo2025octopus,
      title={Octopus: Alleviating Hallucination via Dynamic Contrastive Decoding}, 
      author={Wei Suo and Lijun Zhang and Mengyang Sun and Lin Yuanbo Wu and Peng Wang and Yanning Zhang},
      year={2025},
      eprint={2503.00361},
      archivePrefix={arXiv},
      primaryClass={cs.CV},
      url={https://arxiv.org/abs/2503.00361}, 
}

@inproceedings{wightman2023strength,
  title={Strength in numbers: Estimating confidence of large language models by prompt agreement},
  author={Wightman, Gwenyth Portillo and Delucia, Alexandra and Dredze, Mark},
  booktitle={Proceedings of the 3rd Workshop on Trustworthy Natural Language Processing (TrustNLP 2023)},
  pages={326--362},
  year={2023}
}

@article{woo2024don,
  title={Don't Miss the Forest for the Trees: Attentional Vision Calibration for Large Vision Language Models},
  author={Woo, Sangmin and Kim, Donguk and Jang, Jaehyuk and Choi, Yubin and Kim, Changick},
  journal={arXiv preprint arXiv:2405.17820},
  year={2024}
}

@article{yang2023exploring,
  title={Exploring diverse in-context configurations for image captioning},
  author={Yang, Xu and Wu, Yongliang and Yang, Mingzhuo and Chen, Haokun and Geng, Xin},
  journal={NeurIPS},
  volume={36},
  pages={40924--40943},
  year={2023}
}

@article{peng2025lmm,
  title={Lmm-r1: Empowering 3b lmms with strong reasoning abilities through two-stage rule-based rl},
  author={Peng, Yingzhe and Zhang, Gongrui and Zhang, Miaosen and You, Zhiyuan and Liu, Jie and Zhu, Qipeng and Yang, Kai and Xu, Xingzhong and Geng, Xin and Yang, Xu},
  journal={arXiv preprint arXiv:2503.07536},
  year={2025}
}

@article{wu2025generalization,
  title={On the generalization of sft: A reinforcement learning perspective with reward rectification},
  author={Wu, Yongliang and Zhou, Yizhou and Ziheng, Zhou and Peng, Yingzhe and Ye, Xinyu and Hu, Xinting and Zhu, Wenbo and Qi, Lu and Yang, Ming-Hsuan and Yang, Xu},
  journal={arXiv preprint arXiv:2508.05629},
  year={2025}
}
}

\clearpage
\setcounter{page}{1}
\maketitlesupplementary

\appendix

\section{Appendix}
The following appendix contains supplementary details and experimental results excluded from the main paper due to space constraints. The overall appendix includes: B) adaptive plausibility constraint; C) generation of counterfactual inputs; D) additional experimental results and analyses.



\section{Adaptive Plausibility Constraint}
\label{sec:APC}
As mentioned in the main paper, we adopt adaptive plausibility constraint from VCD~\citep{leng2024mitigating} and M3ID~\citep{favero2024m3id} as a post-processing step before sampling output tokens. This constraint masks tokens with low logit values under the original input, ensuring that low-confidence tokens are not sampled as final outputs. Specifically, the constraint can be formulated as:
\begin{align}
    & Z_{vcd}(v,v^*,q)_k = -\infty, \\
    & \text{s.t. } Z(v,q)_k < \max_k(Z(v,q)) + \log(\beta),
    \label{eq-appendix-constraint1}
\end{align}
where $k$ is the token index for logits; the logit with value $-\infty$ ensures that $p_{vcd}(y|v,v^*,q)_k = 0$ for the masked tokens; $\beta$ is the threshold; $\max_k(Z(v,q))$ is the largest logit value for original inputs.

The rationale behind the Adaptive Plausibility Constraint is that, although the output distribution under the original input may be biased, it can still serve as a valid filter to identify plausible candidate tokens. Only tokens with logits greater than $\max_k(Z(v,q)) + \log(\beta)$ are allowed to receive VCD logits and participate in final sampling. In contrast, low-confidence candidates with insufficient logits are directly masked out. As shown in Table~\ref{tab-appx-constraint}, removing the adaptive plausibility constraint leads to a performance drop for SCI$_5$ on the B/S/BS subsets, and results in an even greater performance degradation on the original datasets as we expected.

For the proposed Self-Critical Inference (SCI) framework, we slightly change the constraint as follows:
\begin{align}
    & p_{\text{SCI}}(y|\bm{v},\bm{q}) = 0, \\
    & \text{s.t. } TC_k/\tau_1 < \max_k(TC/\tau_1) + \log(\beta),
    \label{eq-appendix-constraint2}
\end{align}
where the key difference is that we use Textual Counterfactual (TC) logits, scaled by a temperature factor, to replace the original logits as the masking criterion, as we believe TC provides more consistent predictions. The final output tokens are then sampled from the unmasked candidates with non-zero probabilities.

In our experiments, the default threshold $\beta$ is set to 0.3 following the previous paper~\citep{favero2024m3id} for all DRBench experiments. We consider $\beta$ as a trade-off parameter between relying on de-biased logits and original logits. When $\beta$ approaches 1.0, the final output token closely resembles that produced by the original inputs. In contrast, when $\beta$ approaches 0.0, the constraint becomes negligible, and the output behaves as if no filtering is applied. For experiments on original LVLM datasets, we increase $\beta$ by 0.5 to 0.8, as these datasets exhibit less bias and the outputs are generally closer to those produced by the original inputs.

\begin{table}
    \centering
    \fontsize{6.5}{10}\selectfont
    \begin{tabular}{lccccc} 
    \toprule
    Methods & Constraint & Original & B Subset & S Subset & BS Subset \\
    \midrule
    Qwen2-VL & NA & 81.12 & 6.10 & 37.59 & 15.46\\ 
    Qwen2-VL-SCI$_5$ & \ding{55} & 68.93 & 26.16 & 34.04 & 27.63\\
    Qwen2-VL-SCI$_5$ & \checkmark & 81.03 & 29.65 & 40.43 & 32.55\\
    \bottomrule
    \end{tabular}
    \caption{Ablation study for the adaptive plausibility constraint. To evaluate the effect of adaptive plausibility constraint, we conducted experiments on validation sets of original 6 datasets together with B(ias)/S(ensitive)/BS Subsets.}
    \label{tab-appx-constraint}
\end{table}

\begin{table*}
    \setlength{\tabcolsep}{4pt}
    \centering
    \begin{tabular}{lcccc} 
    \toprule
    Method & Qwen2-VL & Qwen2-VL-SCI$_3$ & Qwen2-VL-SCI$_5$ & Qwen2-VL-SCI$_7$ \\
    \midrule
    Inference Time (w/o batch inference) & 540.47ms & 1599.65ms & 2707.16ms & 3611.18ms\\
    Inference Time (w/ batch inference) & 540.47ms & 697.24ms & 978.14ms & 1342.86ms\\
    \bottomrule
    \end{tabular}
    \caption{We report the average inference time per sample on the MMStar dataset using one A800 GPU to illustrate the computational overhead introduced by SCI. Note that the baseline speed w/o batch inference sequentially conduct each counterfactual inference round, while w/ batch inference, all counterfactual inference rounds are conducted in one batch. Therefore, the later is significantly faster than the baseline speed.}
    \label{tab-ablation-time}
\end{table*}

\section{Generation of Counterfactual Inputs}
\label{sec:cf}
In this section, we provide further details on the generation of counterfactual inputs. For the Visual Counterfactual input VC-Color0, we directly set the RGB values of all pixels in the input image to (0, 0, 0), resulting in a completely black image. For VC-Noise400 and VC-Noise500, we follow the method used in VCD~\citep{leng2024mitigating}, where Gaussian noise is added to simulate the forward diffusion process~\citep{ho2020denoising} at 400 and 500 time steps, respectively. The mathematical formulation of this forward process is as follows:
\begin{equation}
    v_t = \sqrt{\Bar{\alpha}_t} \cdot v_0 + \sqrt{1-\Bar{\alpha}_t} \cdot \epsilon,
\end{equation}
where $v_t$ is the final noise image at at step $t$; $v_0$ is original image; $\epsilon \sim \mathcal{N}(0, 1)$ is random Gaussian noise; $\Bar{\alpha}_t$ is cumulative product. The detailed implementation is available in the official GitHub repository of VCD. 

\begin{figure*}
    \centering
    \includegraphics[width=1.0\linewidth]{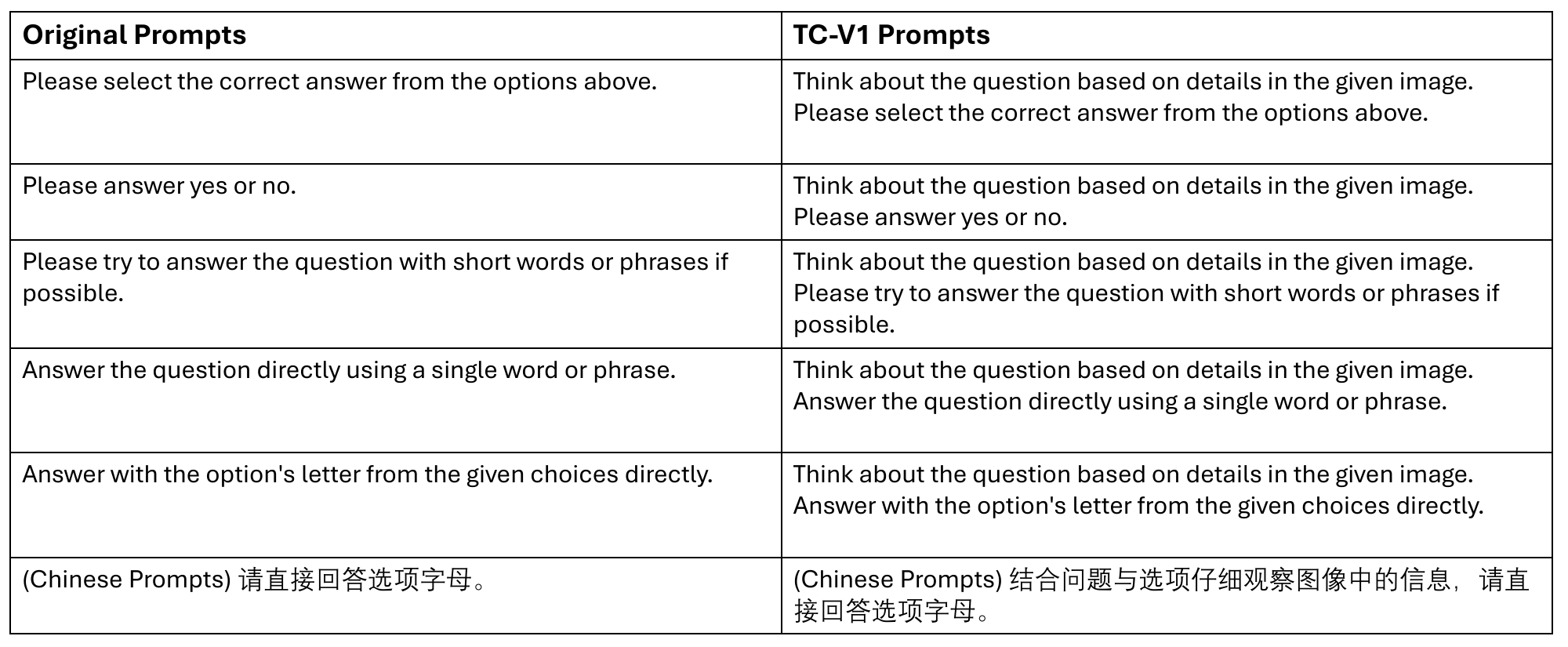}
    \caption{The list of all TC-V1 prompts that add an additional system prompt instructing the model to focus on image details.}
    \label{fig:appx1}
\end{figure*}

\begin{figure*}
    \centering
    \includegraphics[width=1.0\linewidth]{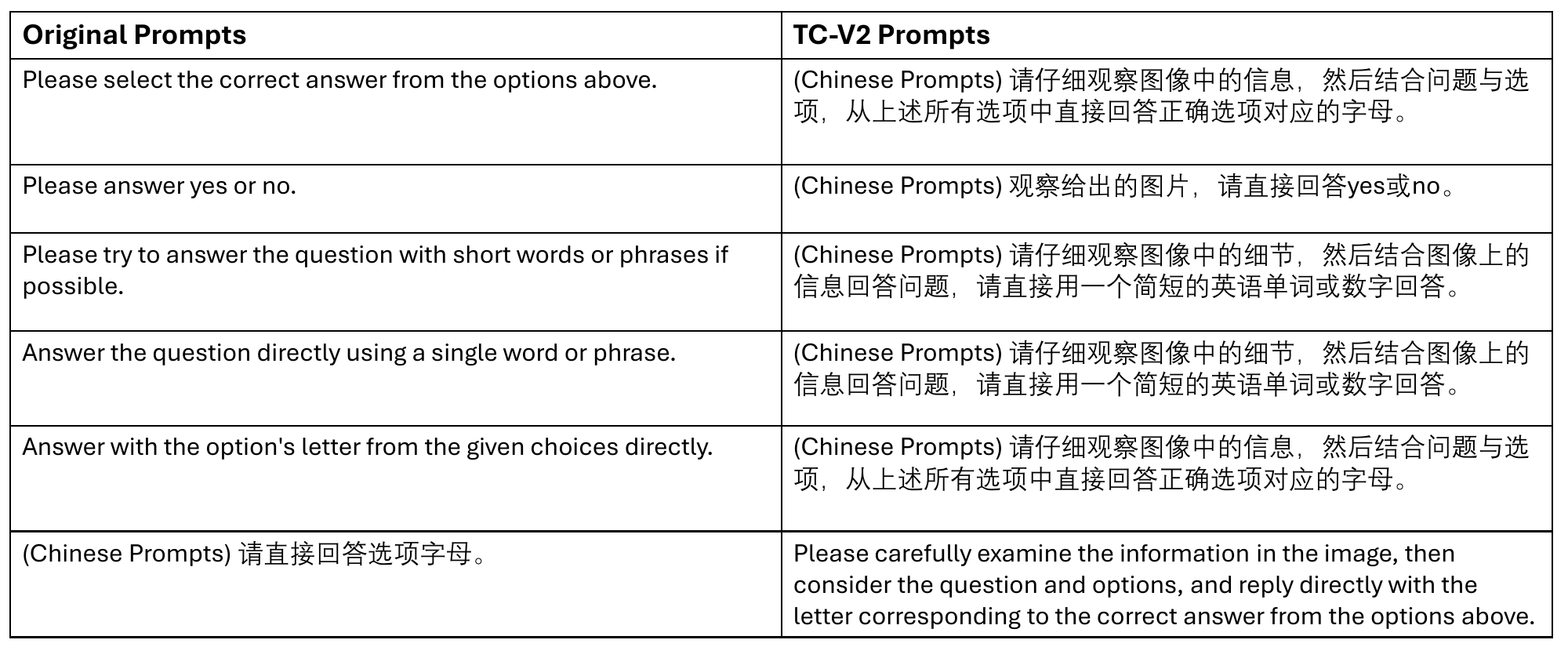}
    \caption{The list of all TC-V2 prompts that further modify the system prompt’s language from English to Chinese or vice versa.}
    \label{fig:appx2}
\end{figure*}

\begin{figure*}
    \centering
    \includegraphics[width=1.0\linewidth]{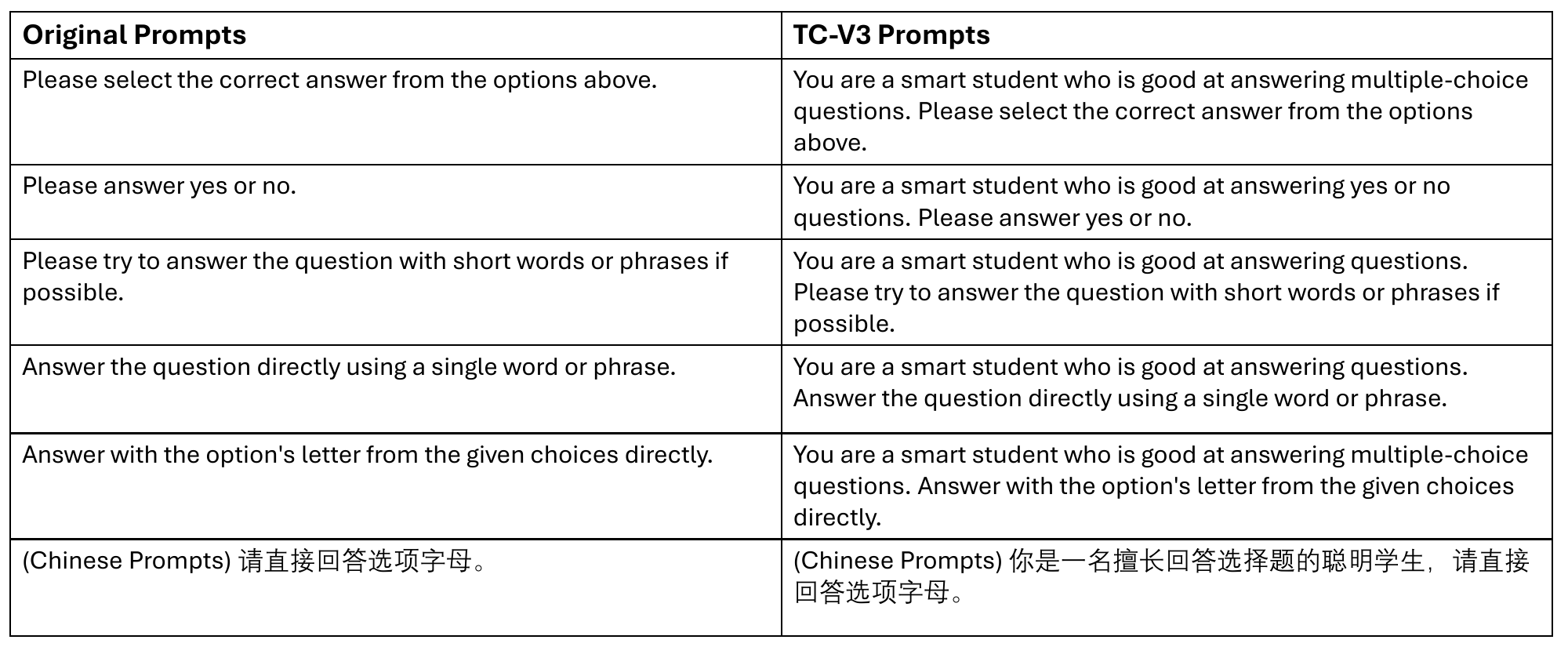}
    \caption{The list of all TC-V3 prompts that inject identity information by prompting the model to respond as a clever student.}
    \label{fig:appx3}
\end{figure*}

For Textual Counterfactual input TC-V1, TC-V2, and TC-V3, as we can see from Figure~\ref{fig:appx1}, Figure~\ref{fig:appx2}, and Figure~\ref{fig:appx3}, each variations provide a semantically equivalent but lexically different prompts. Without change the meaning of instruction, TC-V1 adds an additional system prompt instructing the model to focus on image details, TC-V2  further modifies the system prompt’s language from English to Chinese or vice versa, TC-V3 injects identity information by prompting the model to respond as a clever student.

\section{Additional Experiments}
\label{sec:appexp}
This section will discuss some additional experiments, including ablation studies on hyperparameters, analysis of inference time for SCI, and other supplementary results.

\begin{table}
    \centering
    \fontsize{7.5}{10}\selectfont
    \begin{tabular}{lcccc} 
    \toprule
    Methods & Hyperparameters & MCQ & Others & Overall \\
    \midrule
    Qwen2-VL & - & 11.97 & 22.38 & 15.46\\ 
    Qwen2-VL-SCI$_5$ & $\tau_1=2.0$ $\tau_2=0.2$ & 33.45 & 30.77 & 32.55\\
    \midrule
    Qwen2-VL-SCI$_5$ & $\tau_1=2.0$ $\tau_2=2.0$ & 22.89 & 27.97 & 24.59 \\
    Qwen2-VL-SCI$_5$ & $\tau_1=2.0$ $\tau_2=1.0$ & 26.06 & 29.37 & 27.17 \\
    Qwen2-VL-SCI$_5$ & $\tau_1=2.0$ $\tau_2=0.5$ & 32.39 & 30.77 & 31.85 \\
    \midrule
    Qwen2-VL-SCI$_5$ & $\tau_1=20$ $\tau_2=0.2$ & 3.87 & 18.88 & 8.89 \\
    Qwen2-VL-SCI$_5$ & $\tau_1=10$ $\tau_2=0.2$ & 23.59 & 23.77 & 23.65\\
    Qwen2-VL-SCI$_5$ & $\tau_1=1.0$ $\tau_2=0.2$ & 28.17 & 26.57 & 27.63\\
    Qwen2-VL-SCI$_5$ & $\tau_1=0.2$ $\tau_2=0.2$ & 11.97 & 20.97 & 14.99\\
    \bottomrule
    \end{tabular}
    \caption{Ablation study for temperature scaling hyperparameters $\tau_1$ and $\tau_2$ of SCI. Experiments are conducted under validation set of BS Subset.}
    \label{tab-ablation-b}
\end{table}

\noindent\textbf{Ablation study for hyperparameters.} As shown in Table~\ref{tab-ablation-b}, we select the temperature scaling hyperparameters for the TC and VC logits based on validation performance on the BS Subset. For fair comparison, the hyperparameters were select on SCI$_5$ under base model Qwen2-VL and directly apply to LLaVA-NeXT. The temperature scaling $\tau_2$ for VC is fixed as $0.2$ across SCI$_3$, SCI$_5$, and SCI$_7$, because the logits distribution of VC would not change with the number of visual counterfactual inputs. As to the temperature scaling $\tau_1$ for TC, since the calculation of TC involves maximum cross all outputs using different textual counterfactual inputs, the logits distribution of TC would change with number of textual counterfactual variations. Therefore, we decide to intuitively add $0.5$ to $\tau_1$ to prevent the distribution change when there is one more textual variation added to SCI. 

\begin{table*}
    \setlength{\tabcolsep}{4pt}
    \centering
    \begin{tabular}{lcccccccccccc} 
    \toprule
    \multirow{2}{*}{\textbf{Method}}  & \multirow{2}{*}{}
          & \multicolumn{3}{c}{\textbf{Bias Subset}}  
          &  & \multicolumn{3}{c}{\textbf{Sensitivity Subset}} 
          &  & \multicolumn{3}{c}{\textbf{BS Subset}}  \\ 
    \cmidrule{3-5}\cmidrule{7-9}\cmidrule{11-13}
    & & MCQ & Others & Overall & & MCQ & Others & Overall & & MCQ & Others & Overall \\
    \midrule
    LLaVA-NeXT & & 0.0 & 0.0 & 0.0 & & 39.2 & 37.63 & 38.63 & & 15.91 & 27.58 & 18.75           \\
    LLaVA-NeXT-TCF-V1 & & 3.20 & 6.38 & 3.71 & & 36.62 & 26.80 & 33.02 & & 14.86 & 19.65 & 16.02\\
    LLaVA-NeXT-TCF-V2 & & 5.80 & 8.70 & 6.26 & & 24.08 & 33.51 & 27.54 & & 9.77 & 24.56 & 13.36\\
    LLaVA-NeXT-TCF-V3 & & 3.09 & 3.19 & 3.11 & & 38.61 & 34.54 & 37.11 & & 15.99 & 25.31 & 18.26      \\
    LLaVA-NeXT-VCF-Color0 & & 4.59 & 4.06 & 4.50 & & 27.26 & 23.54 & 25.90 & & 13.85 & 18.01 & 14.86  \\
    LLaVA-NeXT-VCF-Noise400 & & 6.63 & 3.19 & 6.08 & & 27.96 & 23.71 & 26.40 & & 14.98 & 17.51 & 15.60\\
    LLaVA-NeXT-VCF-Noise500 & & 6.30 & 3.48 & 5.85 & & 27.16 & 23.02 & 25.65 & & 14.54 & 17.63 & 15.29\\
    LLaVA-NeXT-TIE & & 12.98 & 23.48 & 14.66 & & 39.00 & 57.56 & 45.81 & & 21.89 & 44.21 & 27.31    \\
    LLaVA-NeXT-VCD & & 12.65 & 25.51 & 14.71 & & 40.50 & 56.53 & 46.38 & & 22.54 & 44.58 & 27.89   \\
    LLaVA-NeXT-M3ID & & 16.91 & 25.22 & 18.24 & & 39.90 & 56.36 & 45.94 & & 24.15 & 44.33 & 29.05  \\
    \cellcolor{lightCyan}LLaVA-NeXT-SCI$_3$ (ours) & \cellcolor{lightCyan} & \cellcolor{lightCyan}21.22 & \cellcolor{lightCyan}35.36 & \cellcolor{lightCyan}23.48 & \cellcolor{lightCyan} & \cellcolor{lightCyan}39.60 & \cellcolor{lightCyan}60.31 &  \cellcolor{lightCyan}47.20 & \cellcolor{lightCyan} & \cellcolor{lightCyan}27.14 & \cellcolor{lightCyan}50.13 & \cellcolor{lightCyan}32.72 \\
    \cellcolor{lightCyan}LLaVA-NeXT-SCI$_5$ (ours) & \cellcolor{lightCyan} &\cellcolor{lightCyan}23.81 & \cellcolor{lightCyan}37.97 & \cellcolor{lightCyan}26.08 & \cellcolor{lightCyan} & \cellcolor{lightCyan}\textbf{40.60} & \cellcolor{lightCyan}\textbf{60.65} & \cellcolor{lightCyan}\textbf{47.95} & \cellcolor{lightCyan} & \cellcolor{lightCyan}28.80 & \cellcolor{lightCyan}51.01 & \cellcolor{lightCyan}34.19 \\
    \cellcolor{lightCyan}LLaVA-NeXT-SCI$_7$ (ours) & \cellcolor{lightCyan} &\cellcolor{lightCyan}\textbf{24.86} & \cellcolor{lightCyan}\textbf{38.26} & \cellcolor{lightCyan}\textbf{27.01} & \cellcolor{lightCyan} & \cellcolor{lightCyan}40.10 & \cellcolor{lightCyan}\textbf{60.65} \cellcolor{lightCyan} & \cellcolor{lightCyan}47.64 & \cellcolor{lightCyan} & \cellcolor{lightCyan}\textbf{29.68} & \cellcolor{lightCyan}\textbf{51.26} & \cellcolor{lightCyan}\textbf{34.92}\\
    \midrule
    Qwen2-VL & & 5.37 & 8.56 & 6.11 & & 38.10 & 34.41 & 36.06 & & 10.78 & 23.59 & 14.52    \\ 
    Qwen2-VL-TCF-V1  & & 6.11 & 11.31 & 7.32 & & 36.51 & 36.01 & 36.23 & & 10.38 & 24.37 & 14.46\\
    Qwen2-VL-TCF-V2 & & 7.59 & 15.90 & 9.52 & & 40.87 & 34.41 & 37.3 & & 12.07 & 23.00 & 15.26\\
    Qwen2-VL-TCF-V3 & & 6.30 & 8.87 & 6.89 & & 37.70 & 34.41 & 35.88 & & 11.02 & 22.42 & 14.35\\
    Qwen2-VL-VCF-Color0 & & 5.83 & 6.73 & 6.04 & & 20.24 & 28.94 & 25.04 & & 8.77 & 18.52 & 11.62\\
    Qwen2-VL-VCF-Noise400 & & 7.59 & 21.41 & 10.80 & & 21.03 & 25.72 & 23.62 & & 10.22 & 24.17 & 14.29 \\
    Qwen2-VL-VCF-Noise500 & & 7.59 & 21.71 & 10.87 & & 20.63 & 27.33 & 24.33 & & 10.62 & 25.15 & 14.86  \\
    Qwen2-VL-TIE & & 16.20 & 16.82 & 16.35 & & 45.63 & 36.66 & 40.67 & & 20.27 & 27.29 & 22.32  \\
    Qwen2-VL-VCD & & 15.74 & 21.71 & 17.13 & & 46.83 & 40.84 & 43.52 & & 20.11 & 30.41 & 23.12 \\
    Qwen2-VL-M3ID & & 19.81 & 21.71 & 20.26 & & \textbf{47.22} & 41.16 & 43.87 & & 23.65 & 30.6 & 25.68 \\
    \cellcolor{lightCyan}Qwen2-VL-SCI$_3$ (ours) \cellcolor{lightCyan} & \cellcolor{lightCyan} & \cellcolor{lightCyan}21.67 & \cellcolor{lightCyan}26.30 & \cellcolor{lightCyan}22.74 & \cellcolor{lightCyan} & \cellcolor{lightCyan}44.05 & \cellcolor{lightCyan}42.44 & \cellcolor{lightCyan}43.16 & \cellcolor{lightCyan} & \cellcolor{lightCyan}24.54 & \cellcolor{lightCyan}32.75 & \cellcolor{lightCyan}26.94 \\
    \cellcolor{lightCyan}Qwen2-VL-SCI$_5$ (ours) \cellcolor{lightCyan} & \cellcolor{lightCyan} & \cellcolor{lightCyan}24.91 & \cellcolor{lightCyan}25.69 & \cellcolor{lightCyan}25.09 & \cellcolor{lightCyan} & \cellcolor{lightCyan}\textbf{47.22} & \cellcolor{lightCyan}42.44 & \cellcolor{lightCyan}44.58 & \cellcolor{lightCyan} & \cellcolor{lightCyan}28.00 & \cellcolor{lightCyan}33.14 & \cellcolor{lightCyan}29.50\\
    \cellcolor{lightCyan}Qwen2-VL-SCI$_7$ (ours) \cellcolor{lightCyan} & \cellcolor{lightCyan} & \cellcolor{lightCyan}\textbf{27.04} & \cellcolor{lightCyan}\textbf{29.66} & \cellcolor{lightCyan}\textbf{27.65} & \cellcolor{lightCyan} & \cellcolor{lightCyan}\textbf{47.22} & \cellcolor{lightCyan}\textbf{45.98} & \cellcolor{lightCyan}\textbf{46.54} & \cellcolor{lightCyan} & \cellcolor{lightCyan}\textbf{29.61} & \cellcolor{lightCyan}\textbf{36.84} & \cellcolor{lightCyan}\textbf{31.72}\\
    \bottomrule
    \end{tabular}
    \caption{The complete experiments on Bias Subset, Sensitivity Subset, and BS Subset of the DRBench across two widely used base LVLMs demonstrate the effectiveness of the proposed SCI framework. \textbf{Bold texts} indicate the best result of each column.}
    \label{tab-appx-subset}
\end{table*}

\begin{table*}
    \centering
    \begin{tabular}{lcccccccccc} 
    \toprule
    \multirow{2}{*}{\textbf{Method}} 
    & \multicolumn{6}{c}{\textbf{Single Dataset}} 
    & & \multicolumn{3}{c}{\textbf{Question Type}} \\
    \cmidrule{2-7}\cmidrule{9-11}
     & MMB-C & MMB-E & MME & CCB & MMS & ViLP & & \textbf{MCQ} & \textbf{Others} & \textbf{Overall} \\ 
    \midrule
    LLaVA-NeXT & 78.0 &	79.72 & 79.57 & 47.0 & 44.75 & 51.53 & & 70.12 & 71.86 & 70.46 \\
    LLaVA-NeXT-TC-V1 & 77.46 & {\color{blue}79.95} & 76.20 & 46.75 & 43.92 & 51.53 & & 69.87 & 69.42 & 69.78 \\
    LLaVA-NeXT-TC-V2 & 77.44 & 77.51 & 78.78 & 46.20 & 42.08 & 50.14 & & 68.68 & 70.90 & 69.12 \\
    LLaVA-NeXT-VC-C0 & 29.97 & 31.85 & 50.29 & 27.02 & 25.08 & 28.47 & & 29.66 & 44.29 & 32.55 \\
    LLaVA-NeXT-VC-N500 & 30.69 & 33.08 & 48.29 & 28.25 & 25.0 & 29.03 & & 30.55 & 42.99 & 33.01 \\
    LLaVA-NeXT-TIE & {\color{blue}78.28} & {\color{blue}80.28} & 77.30 & 45.65 & {\color{blue}46.00} & {\color{blue}53.19} & & {\color{blue}70.36} & 70.68 & 70.42 \\
    LLaVA-NeXT-VCD & {\color{blue}78.38} & {\color{blue}80.28} & 78.09 & 46.63 & {\color{blue}45.00} & {\color{blue}54.31} & & {\color{blue}70.44} & 71.55 & {\color{blue}70.66} \\
    LLaVA-NeXT-M3ID & {\color{blue}78.31} & {\color{blue}80.18} & 78.62 & 45.89 & {\color{blue}45.92} & {\color{blue}54.03} & & {\color{blue}70.36} & 71.86 & {\color{blue}70.66} \\
    \cellcolor{lightCyan}LLaVA-NeXT-SCI$_5$ (ours) & \cellcolor{lightCyan}{\color{blue}78.21} & \cellcolor{lightCyan}{\color{blue}80.08} & \cellcolor{lightCyan}{\color{blue}80.15} & \cellcolor{lightCyan}46.20 & \cellcolor{lightCyan}{\color{blue}45.75} & \cellcolor{lightCyan}{\color{blue}53.06} & \cellcolor{lightCyan} & \cellcolor{lightCyan}{\color{blue}70.32} & \cellcolor{lightCyan}{\color{blue}72.70} & \cellcolor{lightCyan}{\color{blue}70.79} \\
    \midrule
    Qwen2-VL & 85.26 & 86.36 & 87.89 & 73.22 & 59.50 & 56.53 & & 80.91 & 79.27 & 80.58\\ 
    Qwen2-VL-TC-V1 & {\color{blue}85.28} & 86.11 & 87.79 & 73.18 & {\color{blue}59.73} & {\color{blue}58.09} & & 80.84 & {\color{blue}79.63} & {\color{blue}80.60}\\
    Qwen2-VL-TC-V2 & 85.26 & {\color{blue}86.39} & {\color{blue}87.96} & 72.92 & {\color{blue}59.53} & 56.37 & & 80.88 & 79.27 & 80.56\\
    Qwen2-VL-VC-C0 & 34.46 & 35.54 & 50.45 & 25.37 & 27.33 & 24.72 & & 32.66 & 43.38 & 34.77\\
    Qwen2-VL-VC-N500 & 31.33 & 31.82 & 50.13 & 25.43 & 28.50 & 26.81 & & 30.29 & 43.72 & 32.94\\
    Qwen2-VL-TIE & {\color{blue}86.00} & {\color{blue}86.59} & 86.52 & {\color{blue}73.84} & 59.00 & {\color{blue}57.08} & & {\color{blue}81.30} & 78.43 & {\color{blue}80.73}\\
    Qwen2-VL-VCD & {\color{blue}86.05} & {\color{blue}86.56} & 86.41 & {\color{blue}73.77} & {\color{blue}60.08} & {\color{blue}57.92} & & {\color{blue}81.42} & 78.58 & {\color{blue}80.86}\\ 
    Qwen2-VL-M3ID & {\color{blue}85.69} & {\color{blue}86.46} & 86.10 & {\color{blue}73.96} & {\color{blue}59.75} & {\color{blue}57.78} & & {\color{blue}81.25} & 78.31 & {\color{blue}80.67}\\
    \cellcolor{lightCyan}Qwen2-VL-SCI$_5$ (ours) & \cellcolor{lightCyan}{\color{blue}85.97} & \cellcolor{lightCyan}{\color{blue}86.67} & \cellcolor{lightCyan}87.36 & \cellcolor{lightCyan}{\color{blue}73.59} & \cellcolor{lightCyan}{\color{blue}59.92} & \cellcolor{lightCyan}{\color{blue}58.06} & \cellcolor{lightCyan} & \cellcolor{lightCyan}{\color{blue}81.39} & \cellcolor{lightCyan}{\color{blue}79.31} & \cellcolor{lightCyan}{\color{blue}80.98} \\
    \bottomrule
    \end{tabular}
    \caption{Experiments on MMB(ench-Dev)-C/E(N-V11), MME, CCB(ench), MMS(tar), and ViLP including all counterfactual inference results used by SCI$_5$. {\color{blue}Blue texts} indicate an improvement over the baseline.}
    \label{tab-appx-orig}
\end{table*}

\noindent\textbf{Inference time and discussion about acceleration techniques.} As shown in Table~\ref{tab-ablation-time}, we first evaluate the computational overhead of the vanilla implementation (sequential counterfactual inference) of SCI by measuring the average inference time per sample on the validation set of MMStar (Qwen2-VL BS Subset) using a single A800 GPU with Flash Attention 2.7. Specifically, we compare the original inference with SCI$_3$, SCI$_5$, and SCI$_7$. Since the vanilla implementation sequentially executes each counterfactual inference with different input variations, the computational overhead scales approximately linearly, resulting in $2.96 \times$, $5.01 \times$, and $6.68 \times$ the base model's inference time, respectively. We then apply a straightforward acceleration technique, called batching inference to improve the efficiency. Since each counterfactual input variations together with the original input can be executed in the forward pass independently, we can put them into one batch and conduct batch parallel acceleration. The efficiency improvement after applying batch inference is significant, the computational overhead of SCI$_3$, SCI$_5$, and SCI$_7$ become $1.29 \times$, $1.81 \times$, and $2.48 \times$, respectively. In future work, we believe that we can use KV cache sharing to further accelerate the SCI. Since each counterfactual input modifies only either the textual or visual modality, we can exploit shared components to reduce redundant calculations. For example, when the visual input is fixed and only textual prompts vary, we can prefill the visual tokens once and reuse the KV cache across all textual variations. While this approach requires additional engineering effort and potentially model fine-tuning, it offers significant theoretical efficiency gains.

\noindent\textbf{The complete experiments on Bias/Sensitive/BS Subsets.} Due to space constraints, the original paper only presented partial results for the Bias/Sensitive/BS Subsets experiments. The complete results are provided in Table~\ref{tab-appx-subset}. Experiments on all counterfactual inference settings with variant inputs are also included. Although LLaVA-NeXT shows 0.0 accuracy on the Bias Subset, as discussed in the main paper, variants such as LLaVA-NeXT-VCF-Color0, LLaVA-NeXT-VCF-Noise400, and LLaVA-NeXT-VCF-Noise500 may still achieve non-zero performance. This is because the Bias Subset is constructed from the combination of LLaVA-NeXT-VCF-Color0 and LLaVA-NeXT-VCF-Noise500 under our proposed setting. An incorrect prediction from one variant may coincidentally be correct in another (yet, it's still a blind guess), allowing for occasional non-zero accuracies in these counterfactual settings.

\end{document}